\documentclass[lettersize,journal]{IEEEtran}
\usepackage{amsmath,amsfonts}
\usepackage{algorithmic}
\usepackage{algorithm}
\usepackage{array}
\usepackage[caption=false,font=normalsize,labelfont=sf,textfont=sf]{subfig}
\usepackage{textcomp}
\usepackage{stfloats}
\usepackage{url}
\usepackage{verbatim}
\usepackage{graphicx}
\usepackage{cite}
\usepackage{makecell}
\usepackage{multicol}
\usepackage{amsmath,amsfonts}
\usepackage{algorithmic}
\usepackage{algorithm}
\usepackage{array}
\usepackage[caption=false,font=normalsize,labelfont=sf,textfont=sf]{subfig}
\usepackage{textcomp}
\usepackage{stfloats}
\usepackage{url}
\usepackage{verbatim}
\usepackage{graphicx}
\usepackage{cite}
\usepackage{booktabs}
\usepackage{amssymb}  
\usepackage{bbm}
\usepackage{color}
\usepackage{diagbox}
\usepackage{tocbibind}

\newcommand{\removelatexerror}{\let\@latex@error\@gobble}
\begin{document}

\title{Calibrating Bayesian Learning via Regularization, Confidence Minimization, and Selective Inference}

\author{Jiayi Huang, Sangwoo Park,~\IEEEmembership{Member,~IEEE}, and Osvaldo Simeone,~\IEEEmembership{Fellow,~IEEE} \IEEEcompsocitemizethanks{\IEEEcompsocthanksitem The authors are with the King’s Communications, Learning \& Information Processing (KCLIP) lab within the Centre for Intelligent Information Processing Systems (CIIPS), Department of Engineering, King’s College London, London WC2R 2LS, U.K. (e-mail: \{jiayi.3.huang, sangwoo.park, osvaldo.simeone\}@kcl.ac.uk).

The work of J. Huang was supported by the King’s College London and China Scholarship Council for their Joint Full-Scholarship (K-CSC) (grant agreement No. 202206150005). The work of O. Simeone was supported by European Union’s Horizon Europe project CENTRIC (101096379), by the Open Fellowships of the EPSRC (EP/W024101/1), and by the EPSRC project (EP/X011852/1). \protect  \\ 
}}

\maketitle
\vspace{-1cm}
\begin{abstract}
  The application of artificial intelligence (AI) models in fields such as engineering is limited by the known difficulty of quantifying the reliability of an AI's decision. A well-calibrated AI model must correctly report its accuracy on in-distribution (ID) inputs, while also enabling the detection of out-of-distribution (OOD) inputs.  Conventional solutions are effective at addressing only one of these two conflicting requirements. This paper proposes a novel solution that endows Bayesian learning with enhanced ID calibration and OOD detection capabilities by integrating calibration regularization for improved ID performance, confidence minimization for OOD detection, and selective calibration to ensure a synergistic use of calibration regularization and confidence minimization. Selective calibration rejects inputs for which the calibration performance is expected to be insufficient, supporting effective OOD detection, while also ensuring ID calibration. Prior art had only considered these ideas in isolation and for frequentist learning. Numerical results illustrate the trade-offs between ID accuracy, ID calibration, and OOD calibration, showing that the proposed novel Bayesian approach achieves the best ID and OOD  performance compared to existing state-of-the-art approaches, at the cost of rejecting a fraction of the inputs.
\end{abstract}


\section{Introduction}

\subsection{Context and Motivation}

Modern artificial intelligence (AI) models, including deep neural networks and large language models, have achieved great success in many domains, even surpassing human experts in specific tasks \cite{herbold2023large}.  However, it is common to hear concerns voiced by experts on the application of AI for safety-critical fields such as engineering  \cite{ovadia2019can} or health care \cite{esteva2017dermatologist}. These concerns hinge on the known difficulty in quantifying the reliability of an AI's decision, as in the well-reported 
\begin{table}[t]
\centering
\caption{Comparison with state of the art}
\begin{tabular}{@{}cccccc@{}}
\toprule
    reference & $\text{learning}^*$ & \thead{\text{calibration} \\ regularization} & \thead{OOD \\ confidence\\ minimization} & \thead{selective \\ inference}  \\ \midrule
 \thead{\cite{kumar2018trainable}, \\ \cite{bohdal2021meta, yoon2023esd, mukhoti2020calibrating, karandikar2021soft}} & F & \checkmark &  & \\
\hline
 \cite{krishnan2020improving}, \cite{huang2023calibration}  & B & \checkmark &  &  \\
\hline
 \cite{choi2023conservative} & F &  & \checkmark & \\
\hline
 \cite{fisch2022calibrated} & F &  &  & \checkmark\\
\hline
ours  & F, B & \checkmark & \checkmark & \checkmark \\ \bottomrule
\end{tabular}
\flushleft{\footnotesize{* ``F'' $=$ frequentist learning; ``B'' $=$ Bayesian learning.}}
\label{tab:my_label}
\end{table}phenomenon of the ``hallucinations'' of large language models \cite{detommaso2024multicalibration, kumar2023conformal, quach2023conformal}. Indeed, in safety-critical applications, \emph{calibration} -- i.e., the property of a model to know when it does not know -- is arguably just as important as accuracy \cite{zecchin2024forking, lindemann2023safe, ren2023robots}.

A well-calibrated AI model must correctly report its accuracy on \emph{in-distribution} (ID) inputs, i.e., on inputs following the same statistics as training data,  while also enabling the detection of \emph{out-of-distribution} (OOD) inputs, i.e., of inputs that are not covered by the training data distribution. 

A conventional approach to improve ID calibration is the application of Bayesian \emph{ ensembling}. A \emph{Bayesian neural network} (BNN) captures epistemic uncertainty via a distribution over the weights of the model, making it possible to evaluate reliability via the level of consensus between models drawn from the model distribution \cite{ovadia2019can, krishnan2020improving,simeone2022machine}. However, owing to computational limitations and model misspecification, practical ensembling strategies do not necessarily enhance ID calibration \cite{masegosa2020learning, knoblauch2019generalized, wenzel2020good}. Furthermore, improvements in ID calibration often degrade OOD detection performance \cite{ovadia2019can, wald2021calibration, henning2021bayesian}.

Focusing on conventional \emph{frequentist neural networks} (FNNs), prior art has introduced a number of notable methods to separately enhance ID calibration and OOD detection. In particular, first, \emph{calibration regularization} improves ID performance by penalizing excessively confident prediction \cite{kumar2018trainable}. Second, \emph{confidence minimization} enhances OOD detection by exposing the model to OOD samples during training \cite{choi2023conservative}. Finally, \emph{selective calibration} adaptively decides to reject inputs for which the calibration error is estimated to be excessively large \cite{fisch2022calibrated}. This work proposes a novel training strategy that endows BNNs with enhanced ID calibration and OOD detection capabilities by integrating calibration regularization for improved ID performance, confidence minimization for OOD detection, and selective calibration to ensure a synergistic use of calibration regularization and confidence minimization. A table summarizing the comparison of our paper to existing studies can be found in Table~\ref{tab:my_label}.

\begin{figure*} [htb] 
    \centering
    \centerline{\includegraphics[scale=0.3]{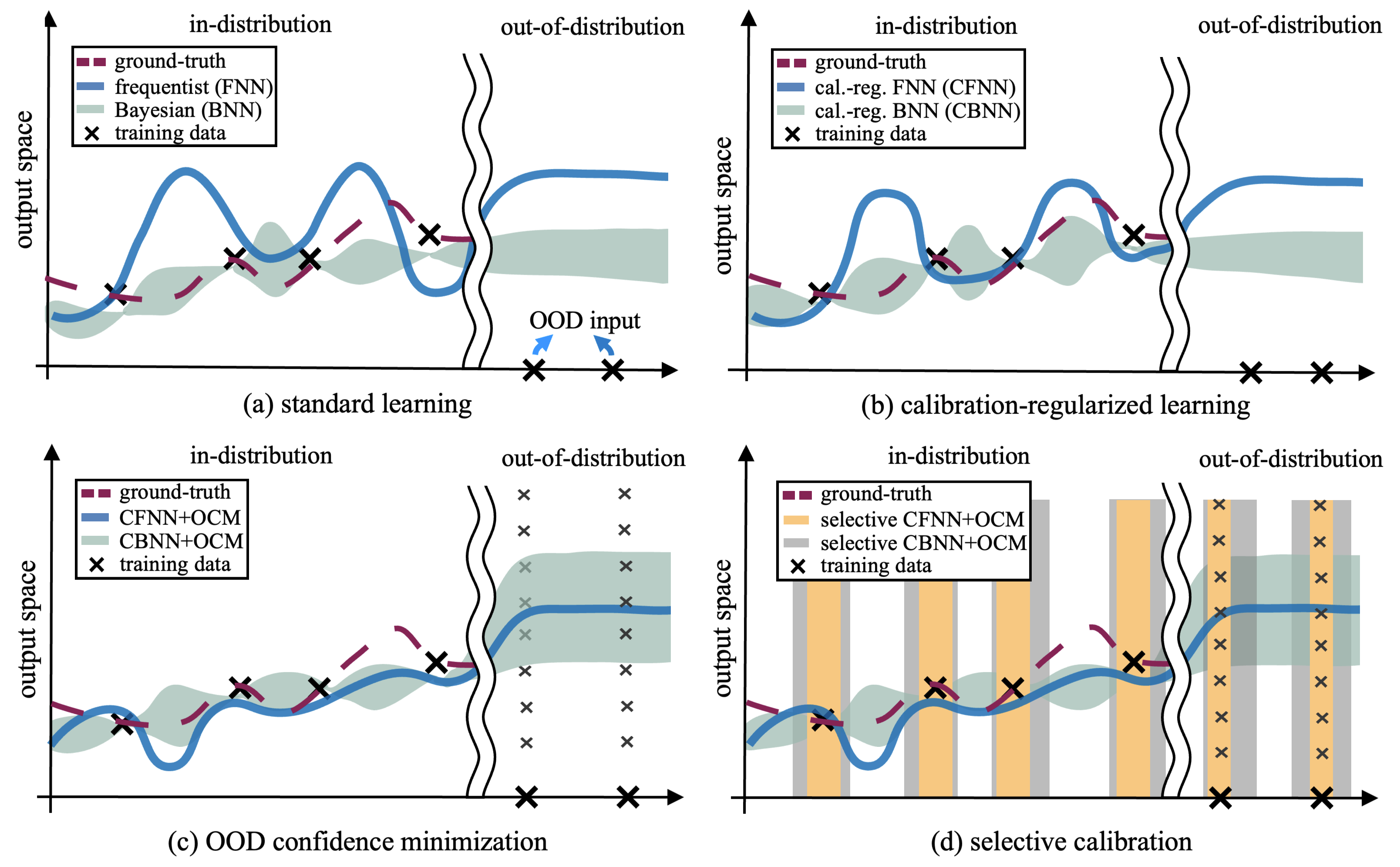}}\vspace{-0.2cm}
        \caption{(a) Standard {frequentist neural networks} (FNNs) generally fail to provide well-calibrated decisions, and improved {in-distribution} (ID) calibration can be achieved via {Bayesian neural networks} (BNNs) \cite{huang2023calibration}. (b) \emph{Calibration regularization} improves ID calibration via a regularizer that penalizes calibration errors \cite{kumar2018trainable}. (c) \emph{Out-of-distribution confidence minimization} (OCM) injects OOD examples during training to improve OOD detection \cite{choi2023conservative}. (d) \emph{Selective calibration} further improves both ID and OOD calibration by only producing decisions for inputs at which uncertainty quantification is deemed to be sufficiently reliable. Prior works \cite{kumar2018trainable, choi2023conservative, fisch2022calibrated} introduced calibration regularization, OCM, and selective calibration as separate methods for FNNs. In contrast, this work presents an integrated training method for BNNs that integrates calibration regularization for improved ID performance, confidence minimization for OOD detection, and selective calibration to ensure a synergistic use of calibration regularization and confidence minimization.}
    \label{fig:6} 
\end{figure*}

\subsection{State of the Art}

As shown in Fig.~\ref{fig:6}(a), conventional FNNs do not capture epistemic uncertainty, i.e., the uncertainty that arises due to limited access to training data. As a result, they often fail to provide a reliable quantification of uncertainty for both ID data -- i.e., in between data points -- and OOD data -- i.e., away from the domain spanned by training data \cite{guo2017calibration, tao2023benchmark}. OOD calibration is measured in terms of the capacity of the model to detect inputs that are too different from available training data, which requires the capacity to measure epistemic uncertainty \cite{henning2021bayesian,tran2022plex}.

In order to enhance the ID calibration of FNNs, \emph{calibration regularization} modifies the training loss by adding measures of ID calibration, such as the \emph{maximum mean calibration error} (MMCE) \cite{kumar2018trainable}, the \emph{ESD} \cite{yoon2023esd}, the \emph{focal loss} \cite{mukhoti2020calibrating}, and the \emph{soft AvUC} \cite{karandikar2021soft}. As shown in Fig.~\ref{fig:6}(b), FNNs with calibration regularization can improve ID calibration, but it does not resolve the issue that FNNs cannot model epistemic uncertainty, as they treat model weights as deterministic quantities, rather than as random variables in BNNs.

As mentioned, improving ID calibration often degrades OOD detection performance, and tailored methods are required to enhance OOD performance  \cite{choi2023conservative, hendrycks2018deep,ovadia2019can, lakshminarayanan2017simple, krishnan2020improving,  tran2022plex}. As illustrated in Fig. 1(c), \emph{out-of-distribution confidence minimization}  (OCM) augments the data set with OOD inputs, which are used to specify a regularizer that favors high uncertainty outside the training set  \cite{choi2023conservative}. However, in so doing, OCM can hurt ID calibration, making the model excessively conservative.

Overall, there is generally a \emph{trade-off between OOD and ID calibration}, and schemes designed to optimize either criterion may hurt the other \cite{choi2023conservative}.

\emph{Selective inference} techniques typically focus on enhancing the accuracy of the model by selecting inputs with the largest model confidence  among the selected inputs \cite{geifman2019selectivenet, huang2020self, pugnana2024deep, fisch2022calibrated}. As a result, these techniques may end up exacerbating issues with ID calibration by favoring over-confidence. In contrast, as shown in Fig.~\ref{fig:6}(d),  \emph{selective calibration} learns how to select inputs that are expected to have a low gap between accuracy and confidence, thus boosting ID calibration \cite{fisch2022calibrated}.

All the papers presented so far have focused on improving either ID or OOD calibration. Furthermore, by focusing on FNNs, these methods are limited in the capacity to provide reliable decisions that fully account for epistemic uncertainty.

\subsection{Main Contributions}

This paper proposes a novel methodology for BNN training that integrates calibration regularization for improved ID performance, OCM for OOD detection, and selective calibration to ensure a synergistic use of calibration regularization and OCM. As illustrated in Fig. 1, this paper constructs this scheme successively by first introducing calibration-regularized Bayesian learning (CBNN); then incorporating OCM to yield CBNN-OCM; and finally integrating also selective calibration to produce the proposed selective CBNN-OCM (SCBNN-OCM). 

Overall, the main contributions are as follows.
\begin{itemize}
    \item We propose CBNN, a novel BNN training scheme that improves the ID calibration performance in the presence of computational complexity constraints and model misspecification by adding a calibration-aware regularizer (see Fig. 1(b)). 
    \item In order to improve OOD detection, CBNN is extended by incorporating an additional regularizer that penalizes confidence on OOD data via OCM. The resulting scheme is referred to as CBNN-OCM (see Fig. 1(c)).  
    \item Since CBNN-OCM can enhance OOD detection performance at the cost of ID performance, we finally propose to further generalize CBNN-OCM via selective calibration (see Fig. 1(d)). The proposed scheme, SCBNN-OCM, selects inputs that are likely to be well calibrated, avoiding inputs whose ID calibration may have been damaged by OCM.
    \item Extensive experimental results on real-world image classification task, including CIFAR-100 data set \cite{krizhevsky2010cifar} and TinyImageNet data set \cite{liang2017principled}, illustrate the trade-offs between ID accuracy, ID calibration, and OOD calibration attained by both FNN and BNN. Among the main conclusions, SCBNN-OCM is seen to achieve best ID and OOD performance as compared to existing state-of-the-art approaches as long as a fraction of inputs is rejected.
\end{itemize}
Versions of CBNN have appeared in \cite{huang2023calibration, krishnan2020improving}, with \cite{huang2023calibration} being an earlier conference version of this work. The authors in \cite{krishnan2020improving} proposed a new calibration-aware regularization term for Bayesian learning, AvUC, whose limitations as a calibration-aware regularizer were demonstrated in \cite{karandikar2021soft}. Reference \cite{huang2023calibration} studied a more general CBNN framework compared to \cite{krishnan2020improving}, while focusing solely on ID calibration performance.

The remainder of the paper is organized as follows. Sec.~\ref{sec: Calibration-Regularized}  summarizes necessary background on ID calibration, calibration-aware training, and Bayesian learning, and it presents CBNN. OOD detection is discussed in Sec.~\ref{sec:OOD}, which introduces CBNN-OCM. Sec.~\ref{sec:general}  describes the selective calibration, and proposes SCBNN-OCM. Sec.~\ref{sec:results} illustrates experimental setting and results. Finally, Sec.~\ref{sec:conclusion} concludes the paper.

\section{Calibration-Regularized Bayesian Learning} \label{sec: Calibration-Regularized}
In this section, we aim at improving the calibration performance of BNNs by introducing a data-dependent regularizer that penalizes calibration errors (see Fig.~\ref{fig:6}(b)). The approach extends the \emph{calibration-regularized FNNs} (CFNNs) of \cite{kumar2018trainable} to CBNNs. We start by defining calibration for classification tasks, followed by a summary of frequentist learning and Bayesian learning. Finally, we propose CBNN.

\subsection{Problem Definition: In-Distribution Calibration} \label{sec: basic concept}

Given a training data set $\mathcal{D}^{\text{tr}} = \{ (x_i, y_i)\}^{|\mathcal{D}^{\text{tr}}|}_{i=1}$, with $i$-th input $x_i \in \mathcal{X}$ and corresponding output $y_i \in \mathcal{Y}$ generated in an independent identically distributed (i.i.d) manner by following an unknown, underlying joint distribution $p(x,y)$, supervised learning for classification optimizes a \emph{probabilistic predictor} $p(y|x, \mathcal{D}^{\text{tr}})$ of output $y$ given input $x$. A probabilistic predictor outputs a \emph{confidence level} $p(y|x, \mathcal{D}^\text{tr})$ for all possible outputs $y \in \mathcal{Y}$ given any input $x \in \mathcal{X}$ and data set $\mathcal{D}^\text{tr}$. Given an input $x$, a \emph{hard} decision $\hat{y}$ can be obtained by choosing the output that has the maximum confidence level, i.e.,  
\begin{align} \label{eq:hard_decision}
    \hat{y} (x) = \arg \max_{y' \in \mathcal{Y}}  p(y'|x,\mathcal{D}^\text{tr}).
\end{align}
Furthermore, the probabilistic predictor $p(y(x)|x, \mathcal{D}^\text{tr})$ provides a measure of the confidence of the prediction for a given input $x$ as
\begin{align} \label{eq:basic_confidence}
    r(x) = p(\hat{y}(x)|x,\mathcal{D}^\text{tr}) = \max_{y' \in \mathcal{Y}} p(y'|x,\mathcal{D}^\text{tr}).
\end{align}

A probabilistic predictor is said to be \emph{perfectly ID calibrated} whenever its average prediction accuracy for any confidence level $r\in [0,1]$ equals $r$. This condition is formalized by the equality
\begin{align} \label{eq:perfect_cal}
    \Pr \left[ y = \hat{y} | p(\hat{y}|x,\mathcal{D}^\text{tr}) = r \right] = r, \text{ for all }  r \in [0,1],
\end{align}
where the probability is taken with respect to the pairs of test data $(x,y)$. For ID calibration, one specifically assume the test pair $(x,y)$ to follow the same distribution $p(x,y)$ of the training data. Under the perfect ID calibration condition (\ref{eq:perfect_cal}), among all input-output pairs $(x,y)$ that have the same confidence level $r$, the average fraction of correct decisions, with $\hat{y} (x) = y$, equals $r$. 

Importantly, perfect calibration does not imply high accuracy \cite{wang2023calibration, tao2023benchmark}. For instance, if the marginal distribution of the labels $y \in \mathcal{Y}$ is uniform, a model that disregards the input $x$, assigning a uniform confidence $p(y|x, \mathcal{D}^{\text{tr}}) = 1/|\mathcal{Y}|$,  achieves perfect calibration, while offering a generally low accuracy.

The extent to which condition (\ref{eq:perfect_cal}) is satisfied is typically evaluated via the \emph{reliability diagram} \cite{degroot1983comparison} and the \emph{expected calibration error} (ECE) \cite{guo2017calibration}. Both approaches estimate the probability in (\ref{eq:perfect_cal}) using a test data set $\mathcal{D}^\text{te}=\{ (x_i^\text{te}, y_i^\text{te}) \}_{i=1}^{|\mathcal{D}^\text{te}|}$, with each pair $(x_i^\text{te}, y_i^\text{te})\sim p(x,y)$ being drawn independently of the training data set $\mathcal{D}^\text{tr}$.

\begin{figure} [t] 
    \centering
    \centerline{\includegraphics[scale=0.3]{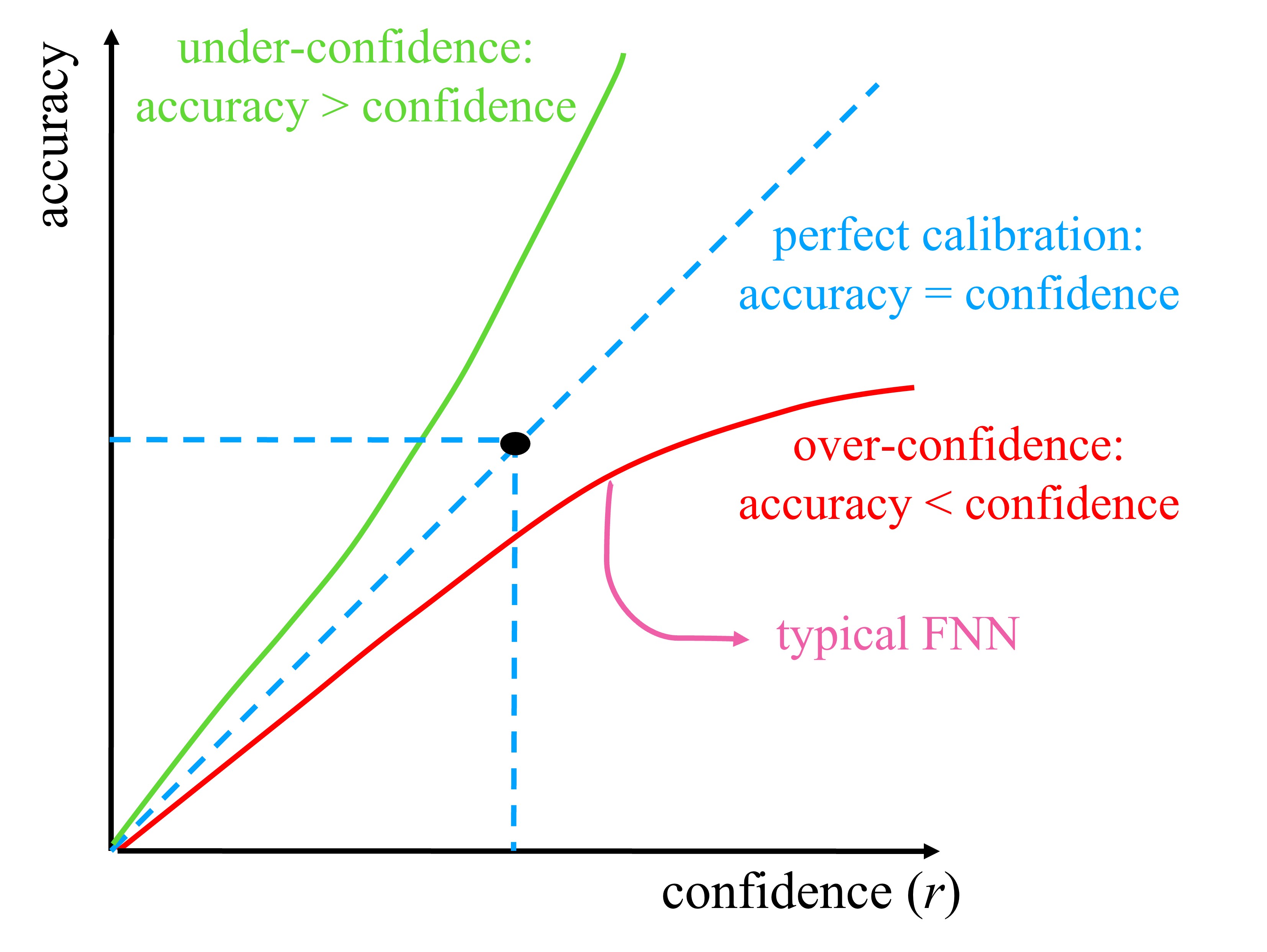}}\vspace{-0.2cm}
    \caption{Reliability diagrams visualize the calibration performance (\ref{eq:perfect_cal}) of the model by evaluating the average accuracy over test examples to which the prediction has the same confidence value $r$. Typically, FNNs return over-confident decisions, for which the accuracy is lower than the confidence obtained by the model.}
    \label{fig:reliability_diagram_illustration} 
\end{figure}

As illustrated in Fig.~\ref{fig:reliability_diagram_illustration}, a \emph{reliability diagram} plots the average confidence produced by the model for all inputs with the same confidence level $r$ in (\ref{eq:perfect_cal}). By (\ref{eq:perfect_cal}), perfect calibration is obtained when the curve is aligned with the diagonal line, i.e., the dashed line in Fig.~\ref{fig:reliability_diagram_illustration}. In contrast, a curve below the diagonal line, e.g., the red line in Fig.~\ref{fig:reliability_diagram_illustration}, indicates over-confidence, while a curve above the diagonal line, such as the green line in Fig.~\ref{fig:reliability_diagram_illustration}, implies an under-confident prediction. 

To elaborate further on calibration measure, denote the \emph{confidence score} associated with the $i$-th test input $x_i^\text{te}$ as 
\begin{align} \label{confidence score}
    r_i^\text{te} = p(\hat{y}_i^\text{te} (x_i^\text{te}) | x_i^\text{te}, \mathcal{D}^\text{tr}), 
\end{align}
given the hard decision (\ref{eq:hard_decision}), and the corresponding \emph{accuracy score} as
\begin{align} \label{correctness score}
    c_i^\text{te} = \mathbbm{1} (\hat{y}_i^\text{te} (x_i^\text{te}) = y_i^\text{te}),
\end{align}
with the indicator function $\mathbbm{1} (\cdot)$ defined as $\mathbbm{1} (\text{true}) = 1$ and $\mathbbm{1} (\text{false}) = 0$. 
To enable an estimate of the reliability diagram, we partition the test data points into \emph{bins} with approximately the same confidence level (\ref{confidence score}). Specifically, the $m$-th bin $\mathcal{B}_m$ contains indices $i$ of the test inputs that have confidence scores lying in the interval $(\frac{m-1}{M}, \frac{m}{M} ]$, i.e., $\mathcal{B}_m = \{i \in \{1,...,M\}: r_i^\text{te} \in ( \frac{m-1}{M}, \frac{m}{M} ] \}$.

Each $m$-th bin is associated with a \emph{per-bin confidence} $\mathrm{conf}(\mathcal{B}_m) ={1}/{|\mathcal{B}_{m}|} \sum_{i \in \mathcal{B}_m} r_i^\text{te}$, and with a \emph{per-bin accuracy} $\mathrm{acc}(\mathcal{B}_m) = {1}/{|\mathcal{B}_{m}|} \sum_{i \in \mathcal{B}_m}  c_i^\text{te}$. An estimate of the reliability diagram is now obtained by plotting $\mathrm{acc}(\mathcal{B}_m)$ as a function of $\mathrm{conf}(\mathcal{B}_m)$ across all bins $m=1,...,M$.

A scalar calibration metric can be extracted from a reliability diagram by evaluating the average discrepancy between the per-bin confidence $\mathrm{conf}(\mathcal{B}_m)$ and the per-bin accuracy $\mathrm{acc}(\mathcal{B}_m)$. Via weighting the contribution of each bin by the corresponding fraction of samples, the ECE is defined as \cite{naeini2015obtaining}
\begin{align} \label{ece}
    \mathrm{ECE} = \sum^{M}_{m=1} \frac{|B_{m}|}{ \sum_{m'=1}^M |B_{m'}|} \left | \mathrm{acc}(B_{m}) - \mathrm{conf}(B_m) \right |.
\end{align}

\subsection{State-of-the-Art: Calibration-Regularized Frequentist Learning} \label{subsection:cfl}

Given a parameterized vector of probabilistic predictor $p(y|x,\theta)$, standard FNN training aims at finding a parameter vector $\theta^\text{FNN}$ that minimizes the training loss $\mathcal{L}(\theta|\mathcal{D}^\text{tr})$. The loss function is typically defined as the negative log-likelihood, or cross-entropy, yielding the solution    
\begin{align} \label{eq:FNN}
    \theta^\text{FNN} = \arg\min_{\theta} \bigg\{ \mathcal{L}(\theta|\mathcal{D}^\text{tr}) = -\sum_{(x,y)\in\mathcal{D}^\text{tr} } \log p(y|x,\theta) \bigg\}.
\end{align} 

Frequentist learning often yields over-confident outputs that fall far short of satisfying the calibration condition (\ref{eq:perfect_cal})  \cite{guo2017calibration, blundell2015weight, lakshminarayanan2017simple}. To address this problem, reference \cite{kumar2018trainable} proposed to modify the training loss by adding a data-dependent \emph{calibration-based} regularizer, namely MMCE, which is defined as 
\begin{align} \label{weighted}
 \mathcal{E}(\theta|\mathcal{D}^\text{tr}) = \Bigg( \sum_{i=1}^{|\mathcal{D}^\text{tr}|} \sum_{j=1}^{|\mathcal{D}^\text{tr}|} \frac{(c_i - {r}_i) (c_j - {r}_j) \kappa({r}_i , {r}_j)}{|\mathcal{D}^\text{tr}|^2} \Bigg)^{\frac{1}{2}},
\end{align}
where $\kappa(\cdot,\cdot)$ is a kernel function. The MMCE (\ref{weighted}) provides a differentiable measure of the extent to which the calibration condition (\ref{eq:perfect_cal}) is satisfied, and the confidence score $r_i$ and correctness score $c_i$ are defined as in (\ref{confidence score}) and (\ref{correctness score}), respectively, with the only difference that they are evaluated based on $i$-th training example $(x_i,y_i)$. Other possible choices for the regulairzer $\mathcal{E}(\theta|\mathcal{D}^\text{tr})$ include the \emph{weighted MMCE} \cite{kumar2018trainable}, the \emph{differentiable ECE} \cite{bohdal2021meta}, the \emph{ESD} \cite{yoon2023esd}, the \emph{focal loss} \cite{mukhoti2020calibrating}, and the \emph{soft AvUC} \cite{karandikar2021soft}.

Equipped with a data-dependent calibration-aware regularizer $\mathcal{E}(\theta|\mathcal{D}^\text{tr})$, CFNN aims at finding a deterministic parameter vector $\theta^\text{CFNN}$ that minimizes the regularized cross-entropy loss as per the optimization \cite{kumar2018trainable}
\begin{align} \label{CA-FNN}
    \theta^\text{CFNN} = \arg\min_{\theta} \big\{\mathcal{L}(\theta|\mathcal{D}^\text{tr}) + \lambda \cdot \mathcal{E}(\theta|\mathcal{D}^\text{tr})\big\}
\end{align}
given a hyperparameter $\lambda>0$. Accordingly, the trained probabilistic predictor is obtained as
\begin{align}
    p(y|x,\mathcal{D}^{\text{tr}}) = p(y|x, \theta^{\text{CFNN}}).
\end{align}
Note that in (\ref{CA-FNN}) the regularizer $\mathcal{E}(\theta|\mathcal{D}^\text{tr})$ depends on the same training data $\mathcal{D}^\text{tr}$ used in the cross-entropy loss $\mathcal{L}(\theta|\mathcal{D}^\text{tr})$. An alternative, investigated in \cite{yoon2023esd}, is to split the training data into two disjoint parts $\mathcal{D}^\text{tr}_1$ and $\mathcal{D}^\text{tr}_2$ with $\mathcal{D}^\text{tr}_1 \cup \mathcal{D}^\text{tr}_2 = \mathcal{D}^\text{tr}$ and evaluate the regularizer using $\mathcal{D}^\text{tr}_2$, yielding the problem $\min_{\theta \in \Theta} \{\mathcal{L}(\theta|\mathcal{D}^\text{tr}_1) + \lambda \cdot \mathcal{E}(\theta|\mathcal{D}^\text{tr}_2)\}$, where the cross-entropy loss is estimated using data set $\mathcal{D}^\text{tr}_1$.

\subsection{Background: Bayesian Learning} \vspace{-0.1cm}
\label{subsubsec:standard_Bayesian}

BNN training leverages \emph{prior} knowledge on the \emph{distribution} $p(\theta)$ of the model parameter vector $\theta$, and it aims at finding a \emph{distribution} $q(\theta)$ over the parameter vector $\theta$ that represents the learner's uncertainty in the model parameter space. The learning objective $\mathcal{F}(q|\mathcal{D}^\text{tr})$, known as \emph{free energy}, accounts for the average loss $\mathbb{E}_{\theta \sim q(\theta) }[\mathcal{L}(\theta|\mathcal{D}^\text{tr})]$, as well for discrepancy between the distribution $q(\theta)$ and prior knowledge $p(\theta)$ as per the sum
\begin{align} \label{free-energy}
    \mathcal{F}(q|\mathcal{D}^\text{tr}) = \mathbb{E}_{\theta \sim q(\theta) }[\mathcal{L}(\theta|\mathcal{D}^\text{tr})] + \beta \cdot \operatorname{KL}(q(\theta) || p(\theta)).
\end{align}
In (\ref{free-energy}), the discrepancy between distribution $q(\theta)$ and prior $p(\theta)$ is captured by the Kullback-Liebler (KL) divergence
\begin{align} \label{kl term}
    \operatorname{KL} (q(\theta) \| p(\theta)) = \mathbb{E}_{q(\theta)} \left[\log \left(\frac{q(\theta)}{p(\theta)}\right) \right],
\end{align}
where we have introduced a hyperparameter $\beta > 0$. The KL term within free energy (\ref{free-energy}) reduces the calibration error by enforcing the adherence of the model parameter distribution $q(\theta)$ to the prior distribution $p(\theta)$.

Accordingly, BNN learning yields the optimized distribution $q^\text{BNN}(\theta)$ by addressing the minimization of the free energy as per
\begin{align} \label{eq:BNN}
    q^\text{BNN}(\theta) = \arg\min_{q(\theta)} \mathcal{F}(q|\mathcal{D}^\text{tr}),
\end{align}
\begin{figure*} [tb] 
    \centering
    \centerline{\includegraphics[scale=0.55]{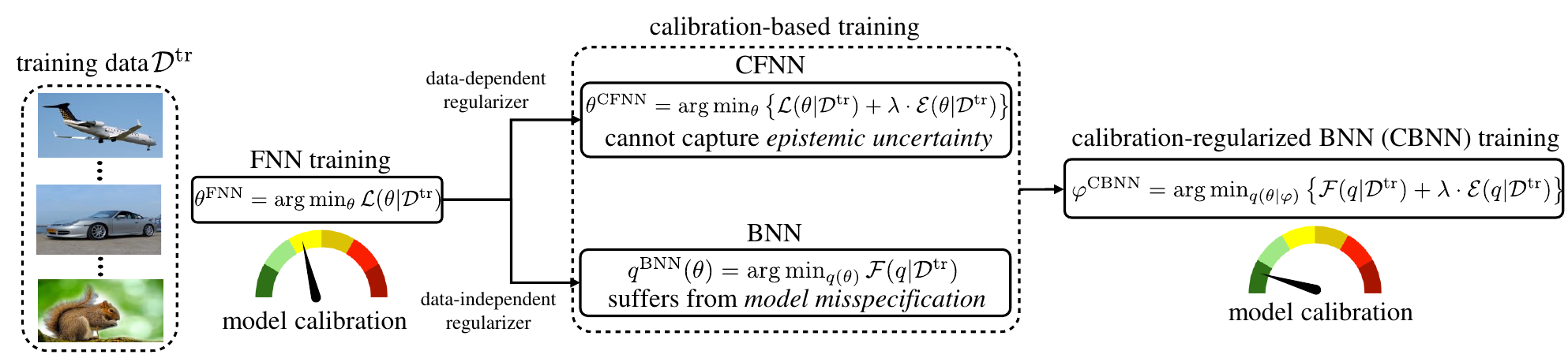}}\vspace{-0.1cm}
    \caption{Standard FNN training \cite{simeone2022machine}  minimizes the cross-entropy loss $\mathcal{L}(\theta|\mathcal{D}^\text{tr})$; CFNN training \cite{kumar2018trainable} minimizes the regularized cross-entropy loss (\ref{CA-FNN}); and BNN learning optimizes the free-energy loss in (\ref{free-energy}) \cite{simeone2022machine}. The proposed CBNN optimizes a regularized free-energy loss with the aim of capturing epistemic uncertainty, like BNNs, while also accounting directly for calibration performance as CFNNs. }
    \label{fig:training_based_cal} 
\end{figure*}and the corresponding trained probabilistic predictor is obtained as the \emph{ensemble} \cite{simeone2022machine}
\begin{align} \label{eq:ensmble}
    p(y|x,\mathcal{D}^\text{tr}) = \mathbb{E}_{ \theta \sim q^\text{BNN}(\theta)} \big[ p(y|x, \theta)\big]. 
\end{align}
Accordingly, the predictive probability distribution (\ref{eq:ensmble}) requires averaging the output distribution $p(y|x,\theta)$ over the model parameter vector $\theta$ sampled from the distribution $q^\text{BNN}(\theta)$.

In practice, the optimization (\ref{eq:BNN}) over distribution $q(\theta)$ is often carried out over a parameterized class of distribution via \emph{variational inference} (VI) \cite{blundell2015weight, gal2016dropout} (see \cite[Ch. 12]{simeone2022machine} for an overview). Specifically, denoting as $q(\theta|\varphi)$ a distribution over parameter vector $\theta$ parameterized by a vector $\varphi$, the problem (\ref{eq:BNN}) is simplified as
\begin{align} \label{eq:simple_BNN}
    \varphi^{\text{BNN}} = \arg \min_{q(\theta|\varphi)} \mathcal{F}(q|\mathcal{D}^{\text{tr}}),
\end{align}
yielding the optimized variational distribution $q^{\text{BNN}}(\theta) = q(\theta|\varphi^{\text{BNN}})$. The parameter vector $\theta$ is typically assumed to follow a Gaussian distribution, with mean $\mu$ and covariance $K$ included in the variational parameters as $\varphi = [\mu, K]$. Furthermore, the covariance $K$ is often constrained to be diagonal with each $i$-th diagonal term modelled as $K_{ii} = \exp(\rho_i)$ for a learnable vector $\rho = [\rho_1,...,\rho_{n_p}]$, where $n_p$ is the size of the model parameter vector $\theta$. For a Gaussian variational distribution, problem (\ref{eq:simple_BNN}) can be addressed via the reparametrization trick \cite{mohamed2020monte}. 

\subsection{Proposed Method: Calibration-Regularized Bayesian Learning} \label{subsec:CB}

In the previous subsections, we have summarized existing ways to improve the ID calibration performance either via a calibration-based, data-dependent, regularizer (Sec.~\ref{subsection:cfl}) or via a prior-based, data-independent, regularizer (Sec.~\ref{subsubsec:standard_Bayesian}). In this subsection, we first describe some limitations of the existing approaches as solutions to improve ID calibration, and then we propose the CBNN framework.

As described in Sec.~\ref{subsection:cfl}, CFNN training disregards epistemic uncertainty, i.e., uncertainty in the model parameter space, by optimizing over a single parameter vector $\theta^\text{CFNN}$ that minimizes a regularized training loss. This may cause the resulting CFNN predictor to remain over-confident on ID test inputs.

By capturing the epistemic uncertainty of the neural networks via ensembling as in (\ref{eq:ensmble}), BNN learning has the potential to yield a better ID-calibrated probabilistic predictor that accounts for model-level {uncertainty}. However, in practice, BNNs may still yield poorly calibrated predictors in the presence of \emph{model misspecification} \cite{masegosa2020learning, morningstar2022pacm, zecchin2023robust}, as well as due to \emph{approximations} such as the parametrization assumed by VI motivated by computational limitations. Model misspecification may refer to settings with a misspecified prior and/or to a misspecified likelihood $p(y|x,\theta)$, which do not reflect well the true data-generating distribution.

In order to handle the limitations of both CFNN and BNN, we propose CBNN. To this end, we first extend the calibration-based regularizer $\mathcal{E} (\theta | \mathcal{D}^{\text{tr}})$, defined in (\ref{weighted}) for a deterministic model parameter vector $\theta$, to a random vector $\theta \sim q(\theta)$ as 
\begin{align} \label{eq:cal_reg_q}
    \mathcal{E} (q | \mathcal{D}^{\text{tr}}) = \mathbb{E}_{\theta \sim q(\theta)}\big[\mathcal{E} (\theta | \mathcal{D}^{\text{tr}})\big].
\end{align}
The key idea underlying CBNN is to leverage the calibration regularizer (\ref{eq:cal_reg_q}) as a way to correct for the ID calibration errors arising from model misspecification and approximate Bayesian inference. Accordingly, adopting VI, CBNN learning aims at finding a variational parameter $\varphi$ by addressing the minimization
    \begin{align} \label{eq:CA-BNN_gen_recipe}
        \varphi^\text{CBNN} = \arg\min_{q(\theta|\varphi)} \big\{\mathcal{F}(q|\mathcal{D}^\text{tr}) + \lambda\cdot \mathcal{E}(q|\mathcal{D}^\text{tr})\big\},
    \end{align} 
for some hyperparameter $\lambda > 0$. In practice, the optimization in (\ref{eq:CA-BNN_gen_recipe}) can be carried out via stochastic gradient descent (SGD) on the variational parameter $\varphi$ via the reparameterization trick, in a manner similar to BNN learning (see Sec.~\ref{subsubsec:standard_Bayesian}).

Finally, the trained probabilistic predictor is obtained as the ensemble
\begin{align}
    p(y|x, \mathcal{D}^{\text{tr}}) = \mathbb{E}_{\theta \sim q(\theta|\varphi^\text{CBNN})}[p(y|x,\theta)].
\end{align}

As anticipated, as compared to existing schemes, CBNNs have two potential advantages. On the one hand, they can improve the calibration performance of CFNNs by taking epistemic uncertainty into account via the use of a variational distribution $q(\theta|\varphi)$ on the model parameter space. On the other hand, they can improve the calibration performance of BNN in the presence of model misspecification and/or approximation errors by modifying the free-energy loss function (\ref{eq:simple_BNN}) through a calibration-driven regularizer. A summary of CBNN with a comparison to FNN, CFNN, and BNN is illustrated in Fig.~\ref{fig:training_based_cal}.

\begin{figure*} [htb] 
    \centering
    \centerline{\includegraphics[scale=0.5]{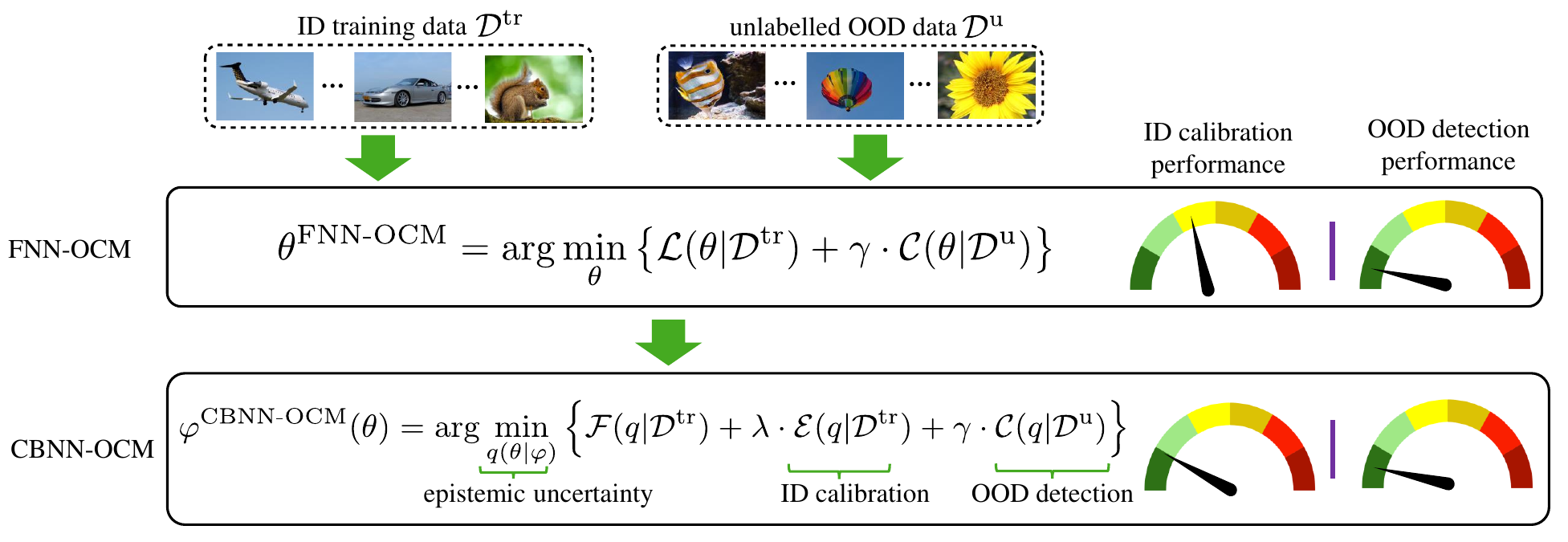}}\vspace{-0.1cm}
    \caption{Unlike standard FNNs that aim at maximizing the accuracy on the ID training data set, FNN-OCM caters to the OOD detection task by maximizing the uncertainty for an unlabelled data set that contains OOD inputs. As a result, the model tends to assign different confidence levels $r(x) = p(\hat{y}(x)|x, \mathcal{D}^\text{tr})$ to ID and OOD inputs. The proposed CBNN-OCM accounts for both ID calibration and OOD detection by capturing epistemic uncertainty as well as OOD uncertainty.}\vspace{-0.1cm}
    \label{fig:OCM} 
\end{figure*}
\section{Bayesian Learning with OOD Confidence Minimization} \label{sec:OOD}

In the previous section, we have designed the CBNN training methodology with the aim at improving ID calibration performance. As discussed in Sec.~I, another important requirement of reliable models is \emph{OOD detection} \cite{tran2022plex, geng2020recent}, i.e., the ability to distinguish OOD data from ID data. In this section, we describe an extension of CBNN that aims at improving the OOD detection performance via the introduction of OCM \cite{choi2023conservative}. To this end, we first describe the OOD detection problem; and then, we review the OCM regularizer presented in \cite{choi2023conservative} to improve OOD detection for FNNs. Finally, we propose the inclusion of an OCM regularizer within the training process of CBNNs, defining the proposed CBNN-OCM scheme (see Fig.~\ref{fig:6}(c)).

\subsection{Problem Definition: OOD Detection}\vspace{-0.1cm}
An important facet of calibration is the capacity of a model to detect OOD inputs, that is, inputs that suffer from a distributional shift with respect to training conditions \cite{tran2022plex}. OOD detection aims at distinguishing OOD inputs from ID inputs by relying on the observation of the confidence levels $p(y|x, \mathcal{D}^\text{tr})$ produced by the model  \cite{ren2019likelihood,hendrycks2016baseline,  choi2023conservative}. The underlying principle is that, if the predictor $p(y|x, \mathcal{D}^\text{tr})$ is well calibrated, OOD inputs should be characterized by a high level of predictive uncertainty, i.e., by a higher-entropy conditional distribution $p(y|x, \mathcal{D}^\text{tr})$, as compared to ID inputs. 

To quantity the effectiveness of OOD detection, one can thus use the discrepancy between the typical confidence levels $r(x) = p(\hat{y}(x)|x, \mathcal{D}^\text{tr})$ in (\ref{eq:basic_confidence}) assigned to the hard decision $\hat{y} (x)$ in (\ref{eq:hard_decision}) for ID inputs and the typical confidence levels $r(x) = p(\hat{y}(x)|x, \mathcal{D}^\text{tr})$ for OOD inputs $x$. Indeed, if the confidence level $r(x)$ tends to be different for ID and OOD inputs, then OOD detection is likely to be successful.

To elaborate, we define $p^{\text{ID}}(r)$ as the distribution of the confidence levels $r(x)$ when $x$ is an ID input, and $p^{\text{OOD}}(r)$ as the distribution of the confidence levels $r(x)$ when $x$ is an OOD input. Then, using standard results on the optimal probability of error for binary detection (see, e.g.,  \cite{polyanskiy2022information}), one can use the \emph{total variation} (TV) distance
\begin{align} \label{tv_distance}
    \text{TV} = \frac{1}{2} \int_{0}^{1} \Big|p^{\text{ID}}(r) - p^{\text{OOD}}(r) \Big| \mathrm{d}r
\end{align}
as a performance measure for OOD detection. In fact, when ID and OOD data are equally likely, the \emph{optimal OOD detection probability} is given by
\begin{align} \label{eq:OOD-detection-probability}
    p_{\text{d}}^{\text{OOD}} = \frac{1}{2}(1+\text{TV}).
\end{align}

\subsection{State-of-the-Art: OOD Confidence Minimization} \label{sub:ocm_fre}
In this subsection, we summarize OCM \cite{choi2023conservative}, a state-of-the-art frequentist learning strategy, which improves the OOD detection performance of FNNs by leveraging an \emph{unlabelled} OOD data set $\mathcal{D}^\text{u}=\{x^\text{u}[i]\}_{i=1}^{|\mathcal{D}^\text{u}|}$. We emphasize that reference \cite{choi2023conservative} concluded that OCM outperforms the other benchmarks on OOD detection, such as MaxLogit \cite{hendrycks2019scaling} and Energy Score \cite{liu2020energy}. The data set $\mathcal{D}^\text{u}$ contains $|\mathcal{D}^\text{u}|$ inputs $x^\text{u}[i]$ that are generated from the marginal OOD input distribution $p_{\text{OOD}}(x)$ \cite{hendrycks2016baseline,  lakshminarayanan2017simple}. Following \cite{choi2023conservative}, the unlabelled data set $\mathcal{D}^\text{u}$ is referred to as the \emph{uncertainty data set}.

Equipped with the uncertainty data set $\mathcal{D}^\text{u}=\{x^\text{u}[i]\}_{i=1}^{|\mathcal{D}^\text{u}|}$, OCM aims at \emph{maximizing} the uncertainty of the predictor when applied to inputs in $\mathcal{D}^\text{u}$. This is done by considering a fictitious labelled data set in which all inputs $x^\text{u}[i]$ appear with all possible labels in set $\mathcal{Y}$. Accordingly, FNN training with OCM, referred to as FNN-OCM, addresses the problem
\begin{align} \label{eq:CM}
    \theta^{\text{FNN-OCM}} = \arg \min_{\theta} \big\{\mathcal{L}(\theta|\mathcal{D}^\text{tr}) + \gamma \cdot \mathcal{C} (\theta | \mathcal{D}^{\text{u}})\big\},
\end{align}
where the \emph{OCM regularization} term is defined as the log-likelihood of the fictitious labelled data set, i.e.,
\begin{align} \label{eq:freq_OCM}
    \mathcal{C}(\theta|\mathcal{D}^\text{u}) = - \sum_{i=1}^{|\mathcal{D}^\text{u}|} \sum_{y\in \mathcal{Y}} \log p(y|x^\text{u}[i],\theta),
\end{align}
given some hyperparameter $\gamma > 0$. It was shown in \cite{choi2023conservative} that fine-tuning the pre-trained frequentist model $p(y|x,\theta^\text{FNN})$, with $\theta^\text{FNN}$ in (\ref{eq:FNN}), by solving problem (\ref{eq:CM}) can significantly aid the OOD detection task. With the obtained solution, the trained probabilistic predictor for FNNs with OCM is obtained as
\begin{align}
    p(y|x,\mathcal{D}^{\text{tr}}) = p(y|x, \theta^{\text{FNN-OCM}}).
\end{align}

\subsection{Proposed Method: Calibration-Regularized Bayesian Learning with OOD Confidence Minimization} \label{subsec:CB-OCM}

Even though FNN-OCM can help separating ID and OOD inputs, it inherits the poor ID calibration performance of FNNs, as it is unable to capture epistemic uncertainty. In contrast, BNNs and CBNNs can enhance ID calibration, while often failing at the OOD detection task due to computational limitations and model misspecification \cite{ovadia2019can, wald2021calibration, henning2021bayesian}. In order to enhance OOD detection, while still benefiting from the advantages of CBNN in ID calibration, in this subsection, we introduce CBNN-OCM by integrating CBNN and OCM. 

To this end, we first generalize the OCM regularizer (\ref{eq:freq_OCM}) so that it can be applied to a random model parameter vector $\theta \sim q(\theta)$ as the average 
\begin{align} \label{eq:infer_CM}
\mathcal{C} (q | \mathcal{D}^{\text{u}}) & =  \mathbb{E}_{\theta \sim q(\theta)}\big[\mathcal{C} (\theta | \mathcal{D}^{\text{u}})\big] \nonumber \\ &= - \mathbb{E}_{\theta \sim q(\theta)} \Bigg[\sum_{i=1 }^{|\mathcal{D}^\text{u}|}  \sum_{y \in \mathcal{Y}} \log p(y|x^\text{u}[i],\theta)\Bigg].
\end{align}
Then, with (\ref{eq:infer_CM}), we propose to modify the objective of CBNN in (\ref{eq:CA-BNN_gen_recipe}) by adding the OCM regularizer as
\begin{align} \label{eq:general_CM_gen_recipe}
\varphi^\text{CBNN-OCM}(\theta) = \arg\min_{q(\theta|\varphi)} \Big \{ \mathcal{F}(q|\mathcal{D}^\text{tr}) & + \lambda \cdot \mathcal{E}(q|\mathcal{D}^\text{tr}) \nonumber \\ &+ \gamma \cdot \mathcal{C}(q|\mathcal{D}^\text{u}) \Big\}. 
\end{align}
Finally, the trained probabilistic predictor is implemented as the ensemble
\begin{align} \label{eq:CBNN-OCM-predictor}
    p(y|x, \mathcal{D}^{\text{tr}}) = \mathbb{E}_{\theta \sim q(\theta|\varphi^\text{CBNN-OCM})}[p(y|x,\theta)].
\end{align}
With the introduction of both data-dependent and data-independent regularizers in (\ref{eq:general_CM_gen_recipe}), CBNN-OCM has the potential to provide reliable ID predictions, while also  facilitating the detection of OOD inputs. A summary of the relation between FNN-OCM and CBNN-OCM is illustrated in Fig.~\ref{fig:OCM}.

\section{Bayesian learning with Selective Calibration} 
\label{sec:general}

While CBNN-OCM is beneficial for OOD detection, it may potentially deteriorate the ID calibration performance of CBNN by modifying the operation of the model on examples that
are hard to identify as ID or OOD. In this section, we introduce \emph{selective calibration} as a way to correct this potential deterioration in ID performance. Selective calibration endows the model with the capacity to reject examples for which the gap between confidence and accuracy is expected to be excessive \cite{fisch2022calibrated}.

Like conventional selective classification \cite{geifman2019selectivenet}, selective calibration adds the option for a model to refuse to produce a decision for some inputs. However, conventional selective classification aims at rejecting examples with expected low accuracy. Accordingly, in order to ensure the successful selection of low-accuracy examples, selective classification requires the underlying model to be well calibrated, though  possibly inaccurate. In contrast, selective calibration accepts examples with both high and low confidence levels, as long as the selector deems the confidence level to be close to the true accuracy. As a result, selective calibration does not require the underlying model to be well calibrated. Rather, it aims at enhancing calibration on the selected examples by making decisions only on inputs for which the model is expected to be well calibrated.

In this section, we first give a brief introduction to selective classification; then, we describe the selective calibration framework proposed in \cite{fisch2022calibrated} for FNNs; and finally we present the SCBNN-OCM framework that combines selective calibration and CBNN-OCM (see Fig.~\ref{fig:6}(d)). \vspace{-0.4cm}

\subsection{Problem Definition: Selective Classification} \vspace{-0.1cm}\label{subsec:sel_cla}

Given a \emph{pre-trained} model $ p(y|x, \mathcal{D}^{\text{tr}})$, conventional selective classification optimizes a \emph{selector} network $g(x|\phi) \in \{0,1\}$ parameterized by a vector $\phi$. An input is accepted if the selector outputs $g(x|\phi) =1$, and is rejected if $g(x|\phi)=0$.

To design the selector, assume that we have access to a held-out \emph{validation data set} $\mathcal{D}^\text{val}=\{ (x_i^\text{val}, y_i^\text{val}) \}_{i=1}^{|\mathcal{D}^\text{val}|}$. Given a selector $g(x|\phi)$, the number of accepted validation examples is $\sum_{x \in \mathcal{D}^\text{val}} g(x|\phi) $. The \emph{selector validation loss} is the loss evaluated only on the accepted samples of data set $\mathcal{D}^{\text{val}}$, i.e.,
\begin{align} \label{eq:sel-cla-loss}
    \mathcal{R}(\phi|\mathcal{D}^{\text{val}}) = - \frac{\sum_{(x,y)\in \mathcal{D}^{\text{val}}} g(x|\phi) \cdot \log p(y|x, \mathcal{D}^{\text{tr}})}{\sum_{(x,y)\in \mathcal{D}^{\text{val}}} g(x|\phi)}.
\end{align}
The optimization of the selector network is formulated in \cite{geifman2019selectivenet} as the minimization of the selective validation loss (\ref{eq:sel-cla-loss}) subject to a target coverage rate $\xi$, with $0 \leq \xi \leq 1$, i.e.,
\begin{align} \label{eq:sel-cla}
    \phi^{\text{S-Cla}} = \arg \min_{\phi} \mathcal{R}(\phi|\mathcal{D}^{\text{val}}) \quad \text{s.t.} \quad \frac{1}{|\mathcal{D}^{\text{val}}|} \sum_{x \in \mathcal{D}^{\text{val}}} g(x|\phi) \geq \xi.
\end{align}

By adopting the design criterion (\ref{eq:sel-cla-loss}), selective classification aims at choosing only the inputs $x$ for pairs $(x,y)$ with large confidence levels $p(y|x, \mathcal{D}^{\text{tr}})$. As anticipated, this objective is sensible if the underlying model $p(y|x, \mathcal{D}^{\text{tr}})$ is sufficiently well calibrated. Reference \cite{geifman2019selectivenet} addressed the constrained optimization in (\ref{eq:sel-cla}) by converting it into an unconstrained problem via a quadratic penalty function (see \cite[Eq. (2)]{geifman2019selectivenet}). 

\subsection{State-of-the-Art: Selective Calibration} \label{sec:background_sel_ca}

Like selective classification, given a pre-trained FNN model $p(y|x, \theta^{\text{FNN}})$, \emph{selective calibration} introduces a selector network $g(x|\phi)$ that reject or accept an input $x$ by setting $g(x|\phi) =0$ or $g(x|\phi) =1$, respectively. However, rather than targeting high-accuracy examples, selective calibration aims at accepting inputs $x$ for which the confidence level $r(x) = p(\hat{y}(x)|x,\theta^\text{FNN})$ is expected to match the true accuracy of the hard decision $\hat{y}(x)$. Accordingly, the goal of the selector is to ensure the calibration condition (\ref{eq:perfect_cal}) when conditioning on the accepted examples. The resulting \emph{perfect selective ID calibration} condition can be formalized as 
\begin{align} \label{eq:perfect_cal_sel}
    \Pr \Big[ y = \hat{y} | p(\hat{y}(x)|x,\theta^\text{FNN} ) = r, & g(x|\phi) = 1 \Big] = r, \nonumber \\ & \text{ for all }  r \in [0,1].
\end{align}
The condition (\ref{eq:perfect_cal_sel}) is less strict than the perfect ID calibration condition (\ref{eq:perfect_cal}), since it is limited to accepted inputs.

\subsubsection{Training the Selector}

Generalizing the MMCE in (\ref{weighted}), the ECE corresponding to the calibration criterion (\ref{eq:perfect_cal_sel}) can be estimated by using the accepted examples within the validation data set. 
This yields the \emph{selective MMCE} (\ref{eq:sel-cal-loss})
\begin{figure*}[tb] 
    \begin{align}  \label{eq:sel-cal-loss}
    &\mathcal{E}^{\text{S-Cal}}(\phi|\mathcal{D}^{\text{val}}) = \left( \frac{\sum_{i=1}^{|\mathcal{D}^\text{val}|} \sum_{j=1}^{|\mathcal{D}^\text{val}|}(c_i - {r}_i) (c_j - {r}_j) g(x_i^{\text{val}}|\phi)g(x_j^{\text{val}}|\phi) \kappa({r}_i , {r}_j)}{ \sum_{i=1}^{|\mathcal{D}^\text{val}|} \sum_{j=1}^{|\mathcal{D}^\text{val}|} g(x_i^{\text{val}}|\phi)g(x_j^{\text{val}}|\phi)}   \right)^{\frac{1}{2}},
    \end{align}
\end{figure*}
with confidence scores $r_i = p(\hat{y}_i^\text{val} (x_i^\text{val}) | x_i^\text{val}, \theta^\text{FNN})$ and correctness scores $c_i = \mathbbm{1} (\hat{y}_i^\text{val} (x_i^\text{val}) = y_i^\text{val})$. Note that the selective MMCE recovers to the MMCE (\ref{weighted}) when the selector accepts all examples, i.e., when $g(x|\phi)=1$ for all inputs $x$. The parameter vector $\phi$ of the selector is obtained by optimizing the selective MMCE (\ref{eq:sel-cal-loss}) by considering the targeted coverage rate constraint $\xi$, with $0 \leq \xi \leq 1$, as
\begin{align} \label{eq:sel-cal}
    \phi^{\text{S-Cal}} = \arg \min_{\phi}  \mathcal{E}^{\text{S-Cal}}(\phi|\mathcal{D}^{\text{val}}) \quad \text{s.t. } \frac{1}{|\mathcal{D}^{\text{val}}|} \sum_{x \in \mathcal{D}^{\text{val}}} g(x|\phi) \geq \xi.
\end{align}

Reference \cite{fisch2022calibrated} addressed the constrained optimization in (\ref{eq:sel-cal}) by turning it into an unconstrained problem via the introduction of regularizer that accounts for the fraction of selected examples. To this end, the discrete-output function $g(x|\phi)$ is relaxed via a continuous-output function $\tilde{g}(r(x|\theta^\text{FNN}), s(x|\theta^\text{FNN})| \phi)$ that takes values within the interval [0,1]. The function $\tilde{g}(r(x|\theta^\text{FNN}), s(x|\theta^\text{FNN})| \phi)$ takes the confidence score
\begin{align}
    r(x|\theta^\text{FNN}) = p(\hat{y}(x)|x,\theta^\text{FNN})
\end{align}
as input, as well as the \emph{non-parametric outlier score vector} $s(x|\theta^\text{FNN})$ which will be specified shortly. All in all, problem (\ref{eq:sel-cal}) is approximately addressed by solving the problem
\begin{align} \label{eq:calibration_selector}
    \phi^{\text{S-Cal}} &= \arg \min_{\phi}  \Bigg\{\tilde{\mathcal{E}}^{\text{S-Cal}} (\phi|\mathcal{D}^{\text{val}}) \nonumber \\ & - \eta \cdot \sum_{x\in\mathcal{D}^\text{val}} \log \left( \tilde{g}(r(x|\theta^\text{FNN}), s(x|\theta^\text{FNN})| \phi) \right)   \Bigg\},
\end{align}
where $\tilde{\mathcal{E}}^{\text{S-Cal}} (\phi|\mathcal{D}^{\text{val}})$, defined in (\ref{eq:tidle_E}), 
\begin{figure*} [htb]
    \begin{align} \label{eq:tidle_E}
    \tilde{\mathcal{E}}^{\text{S-Cal}}(\phi|\mathcal{D}^{\text{val}})  = \left( \sum_{i=1}^{|\mathcal{D}^\text{val}|} \sum_{j=1}^{|\mathcal{D}^\text{val}|}(c_i - {r}_i) (c_j - {r}_j) \tilde{g}(r_i, s_i| \phi) \tilde{g}(r_j, s_j| \phi) \kappa({r}_i , {r}_j)\right)^{\frac{1}{2}}, \\ \hline \nonumber
    \end{align}
\end{figure*} 
uses confidence score $r_i = r(x_i^{\text{val}}|\theta^\text{FNN})$ and non-parametric outlier score vector $s_i = s(x_i^{\text{val}}|\theta^\text{FNN})$, with a coefficient $\eta \geq 0$  that balances selective calibration performance and coverage. Note that function (\ref{eq:tidle_E}) is the numerator of (\ref{eq:sel-cal-loss}) with the relaxed selector $\tilde{g}(r(x|\theta^\text{FNN}), s(x|\theta^\text{FNN})| \phi)$ in lieu of $g(x|\phi)$, and that $-\log(\cdot)$ plays the role of a barrier function to enforce the coverage rate constraint \cite{boyd2004convex}.

\subsubsection{Non-Parametric Outlier Score Vector}  \label{subsubsec:outlier_score_vector}
The non-parametric outlier score vector $s(x|\theta^\text{FNN})$ of an input $x$ has the goal of quantifying the extent to which input $x$ conforms with the input in the training data set $\mathcal{X}^\text{tr} = \{ x_i  \}_{i=1}^{|\mathcal{D}^\text{tr}|}$. The rationale for adding $s(x|\theta^\text{FNN})$ as an input to the selector is that inputs $x$ that are too ``far'' from the training set may yield more poorly calibrated decisions by the model $p(y|x, \theta^\text{FNN})$. The authors of \cite{fisch2022calibrated} included four different features in vector $s(x|\theta^\text{FNN})$, which are obtained from a kernel density estimator, an isolation forest, a one-class support vector machine, and a $k$-nearest neighbor distance, constructing a $4\times1$ score vector $s(x|\theta^\text{FNN})$. In particular, the first element of vector $s(x|\theta^\text{FNN})$ is obtained by the kernel density estimator \cite{scott2015multivariate}
\begin{align} \label{eq:KDE}
    s_1 (x|\theta^\text{FNN}) = \frac{1}{|\mathcal{X}^{\text{tr}}|} \sum_{i=1}^{|\mathcal{X}^{\text{tr}}|} \kappa \left( || z-z^{\text{tr}}_{i} || \right),
\end{align}
where $\kappa(\cdot)$ is a kernel function; $z$ is the output of the last hidden layer of the parametrized network $p(y|x, \theta^\text{FNN})$; and, similarly, $z_i^{\text{tr}}$ is the corresponding feature vector for $p(y|x_i, \theta^\text{FNN})$. Details regarding the other elements of the score vector $s(x|\theta^\text{FNN})$ can be found in Appendix in the supplementary material.

\subsubsection{Inference with the Selector} \label{subsubsec:controlling_coverage}
During testing, the binary selector $g(x|\phi^\text{S-Cal})$ is obtained via thresholding as
\begin{align} \label{eq:cont_to_discrete_sel_cal}
    g(x|\phi^{\text{S-Cal}}) = \mathbbm{1} \left( \tilde{g}(r(x|\theta^\text{FNN}), s(x|\theta^\text{FNN})| \phi^{\text{S-Cal}}) \geq \tau \right),
\end{align}
where $\tau$ is a hyperparameter controlling the fraction of selected examples. In this regard, we note that the targeted coverage rate $\xi$ can be in principle met by changing only the hyperparameter $\eta$ in (\ref{eq:calibration_selector}) without requiring the choice of the threshold $\tau$ in (\ref{eq:cont_to_discrete_sel_cal}). However, this would require retraining of the selector any time the desired coverage level $\xi$ is changed. Hence, reference \cite{fisch2022calibrated} introduced the additional hyperparameter $\tau$ that can adjust the coverage given a fixed, trained, selector $\phi^\text{S-Cal}$. 

\subsection{Proposed Method: Bayesian Learning with Selective Calibration}

In order to extend selective calibration to CBNN-OCM, we propose to apply the selective MMCE criterion in (\ref{eq:sel-cal-loss}) with confidence score and outlier score vector averaged over the model distribution $q(\theta| \varphi^\text{CBNN-OCM})$ in (\ref{eq:general_CM_gen_recipe}). Specifically, the average confidence score $\Bar{r}(x|\varphi^\text{CBNN-OCM})$ is defined as 
\begin{align} \label{eq:r-bar}
    \Bar{r}(x|\varphi^\text{CBNN-OCM}) = \max_{y' \in\mathcal{Y}} \mathbb{E}_{\theta \sim q(\theta|\varphi^\text{CBNN-OCM})} \left[ p(y'|x,\theta) \right],
\end{align}
\begin{figure*} [htb] 
    \centering
    \centerline{\includegraphics[scale=0.5]{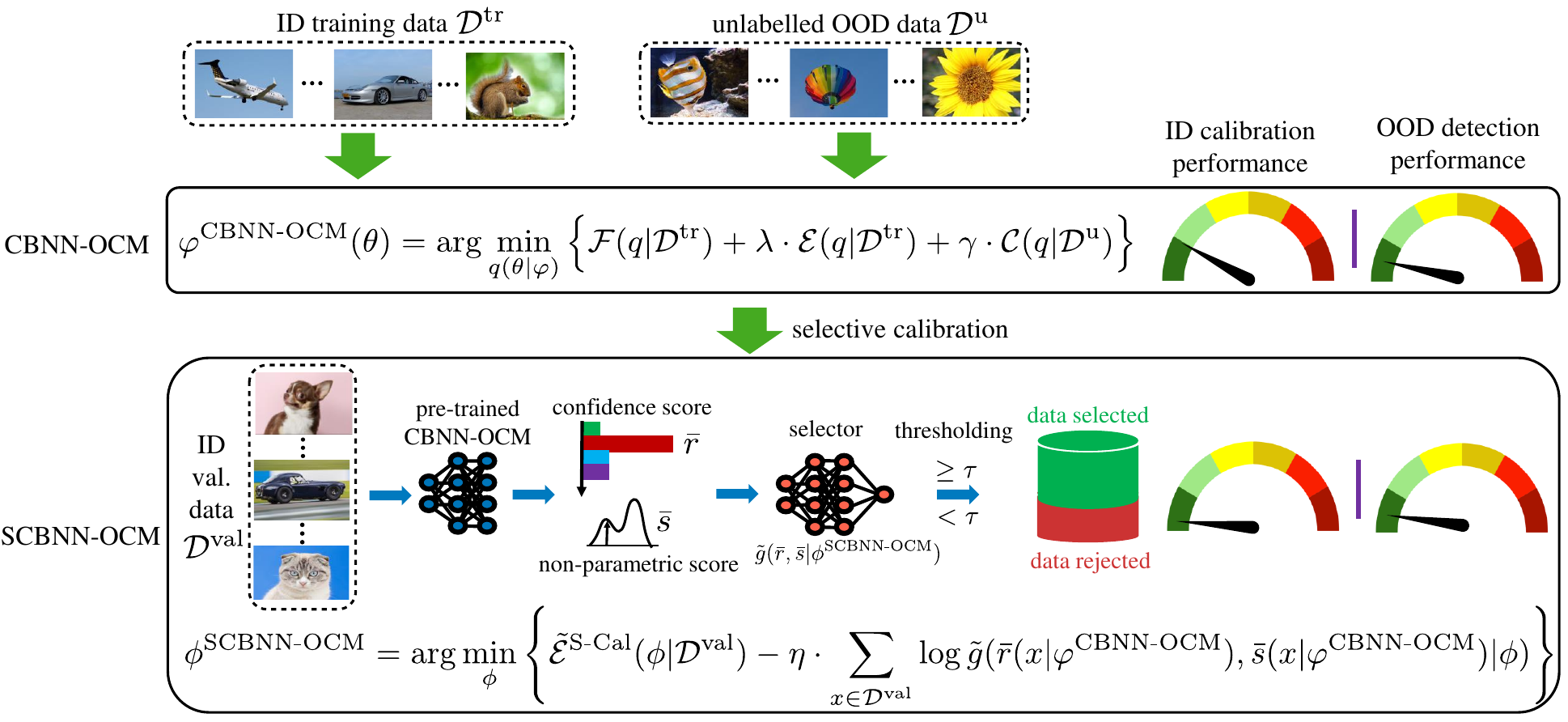}}
    \caption{Given a fixed, pre-trained model parameter vector $\theta \sim q(\theta|\varphi^\text{CBNN-OCM})$, selective calibration aims at achieving well-calibrated decisions (\ref{eq:perfect_cal}) on the selected inputs (hence aiming at   (\ref{eq:perfect_cal_sel})) by rejecting inputs on which the discrepancy between confidence and accuracy is expected to be large.} 
    \label{fig:CBS_CM} 
\end{figure*}and the average non-parametric outlier score vector is given by
\begin{align} \label{eq:s-bar}
    \Bar{s}(x|\varphi^\text{CBNN-OCM}) =  \mathbb{E}_{\theta \sim q(\theta|\varphi^\text{CBNN-OCM})} \left[ s(x|\theta) \right],
\end{align}
with vector $s(x|\theta)$ defined as in the previous subsection.

With these definitions, the selector parameter vector $\phi^\text{SCBNN-OCM}$ is obtained by addressing the problem 
\begin{align} \label{eq:selector_gen_gen_recipe}
    &\phi^{\text{SCBNN-OCM}}  = \arg \min_{\phi}  \Bigg\{  \tilde{\mathcal{E}}^{\text{S-Cal}} (\phi|\mathcal{D}^{\text{val}}) \\ & - \eta \cdot \sum_{x\in\mathcal{D}^\text{val}} \log \tilde{g}(\Bar{r}(x|\varphi^\text{CBNN-OCM}), \Bar{s}(x|\varphi^\text{CBNN-OCM})| \phi)  \Bigg\}, \nonumber
\end{align}
where $\tilde{\mathcal{E}}^{\text{S-Cal}} (\phi|\mathcal{D}^{\text{val}}) $, defined in (\ref{eq:sel-cal-loss-bayes}), 
\begin{figure*}[b] 
    \begin{align} \label{eq:sel-cal-loss-bayes}
        \nonumber \\ \hline \nonumber \\
        \tilde{\mathcal{E}}^{\text{S-Cal}} (\phi|\mathcal{D}^{\text{val}}) = \left( \sum_{i=1}^{|\mathcal{D}^\text{val}|} \sum_{j=1}^{|\mathcal{D}^\text{val}|}(c_i - {\Bar{r}}_i) (c_j - {\Bar{r}}_j) \tilde{g}(\Bar{r}_i, \Bar{s}_i| \phi) \tilde{g}(\Bar{r}_j, \Bar{s}_j| \phi) \kappa({\Bar{r}}_i , {\Bar{r}}_j)    \right)^{\frac{1}{2}},
    \end{align}
\end{figure*}
uses average confidence score $\Bar{r}_i = \Bar{r}(x_i^{\text{val}}|\varphi^\text{CBNN-OCM})$ and average non-parametric outlier score vector $\Bar{s}_i = \Bar{s}(x_i^{\text{val}}|\varphi^\text{CBNN-OCM})$.

Finally, during inference, the binary selector $g(x|\phi^{\text{SCBNN-OCM}})$ is obtained via the thresholding in a manner similar to (\ref{eq:cont_to_discrete_sel_cal}), i.e., 
\begin{align} \label{eq:scbnn-ocm}
    & g(x|\phi^{\text{SCBNN-OCM}})  \\ & = \mathbbm{1} \left( \tilde{g}(\Bar{r}(x|\varphi^\text{CBNN-OCM}), \Bar{s}(x|\varphi^\text{CBNN-OCM})| \phi^{\text{SCBNN-OCM}}) \geq \tau \right), \nonumber
\end{align}
for some threshold $\tau$. A summary of SCBNN-OCM is illustrated in Fig.~\ref{fig:CBS_CM}.

\section{Experiments} \label{sec:results}
In this section, we report on the effectiveness of calibration-based training, OCM, and selective calibration for frequentist and Bayesian learning in terms of ID and OOD calibration performance. Additionally, due to the page limit, the ablation study and backbone study can be checked in the supplementary material.\vspace{-0.3cm}
\vspace{-0.2cm}
\subsection{Setting and Metrics} \label{subsec:setting_and_metrics}
All experiments use the CIFAR-100 data set \cite{krizhevsky2010cifar} to generate ID samples and the TinyImageNet (resized) data set \cite{liang2017principled} to generate OOD samples \cite{choi2023conservative}. The parameterized predictor $p(y|x,\theta)$ adopts the WideResNet-40-2 architecture \cite{zagoruyko2016wide}, and the variational distribution $q(\theta|\varphi)$ is Gaussian with mean and diagonal covariance being the variational parameter $\varphi$. For the selective inference task, we use a 3-layer ReLU activated feed-forward neural network $g(x|\phi)$ with 64 dimensional hidden layer \cite{fisch2022calibrated}.

We consider the following evaluation criteria: (\emph{i}) the \emph{reliability diagram}, which plots accuracy against confidence as defined in Sec.~\ref{sec: basic concept}  with number of bins $M=15$ (we only \begin{figure*} [htb] 
    \centering
    \centerline{\includegraphics[width=\textwidth]{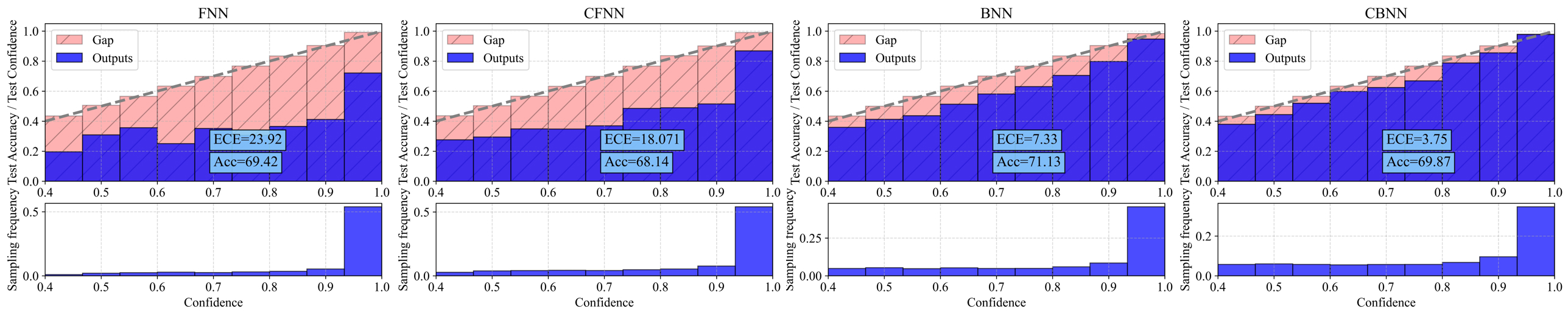}}
    \caption{Reliability diagrams for the CIFAR-100 classification task given the predictor trained using (\emph{i}) FNN (left); (\emph{ii}) CFNN (benchmark) (middle-left); (\emph{iii}) BNN (middle-right); and (\emph{iv}) CBNN (ours) (right).}
    \label{result:CA-RD} 
\end{figure*}visualize the bins that have number of samples no smaller than 100 to avoid undesired statistical noise \cite{ raviv2023modular}); (\emph{ii}) the \emph{ECE}, which evaluates the average discrepancy between per-bin accuracy and per-bin confidence as defined in (\ref{ece}); (\emph{iii}) \emph{ID accuracy}, marked as ``Acc'', which is the proportion of the ID data for which a correct prediction is made; (\emph{iv}) \emph{OOD detection probability}, which is the optimal probability of successfully detecting OOD data, as defined in (\ref{eq:OOD-detection-probability}); and (\emph{v}) \emph{ID coverage rate}, which is the fraction of selected data points on the ID test data set for selective calibration. We refer to Appendix in supplementary document for all the experimental details\footnote{Code can be found at \url{https://github.com/kclip/Calibrating-Bayesian-Learning}.}.\vspace{-0.2cm}

\subsection{Can Calibration-Regularization Enhance ID Calibration for Bayesian Learning?}

We first evaluate the ID calibration performance for (\emph{i}) FNN, defined as in (\ref{eq:FNN}); (\emph{ii}) CFNN with calibration-based regularization \cite{kumar2018trainable}, as defined in (\ref{CA-FNN}); (\emph{iii}) BNN as per (\ref{eq:BNN}); and (\emph{iv}) the proposed CBNN as per (\ref{eq:CA-BNN_gen_recipe}). To this end, we present reliability diagrams, as described in Sec.~\ref{sec: Calibration-Regularized}, as well as the trade-off between accuracy and calibration obtained by varying the hyperparameter $\lambda$ in the designed objectives (\ref{CA-FNN}) and (\ref{eq:CA-BNN_gen_recipe}).

To start, in Fig.~\ref{result:CA-RD}, we show the reliability diagrams for all the four schemes, which are evaluated on the CIFAR-100 data set. Note that the figure also plots test accuracy, reported as ``Acc'', and ECE. BNN is observed to yield better calibrated predictions than FNN with a slight improvement in accuracy, while calibration-based regularization improves the calibration performance for both FNN and BNN. In particular, the ECE is decreased by more than 20\% for frequentist learning and by 50\% for Bayesian learning.
\begin{figure} [htb] 
    \centering
    \centerline{\includegraphics[scale=0.35]{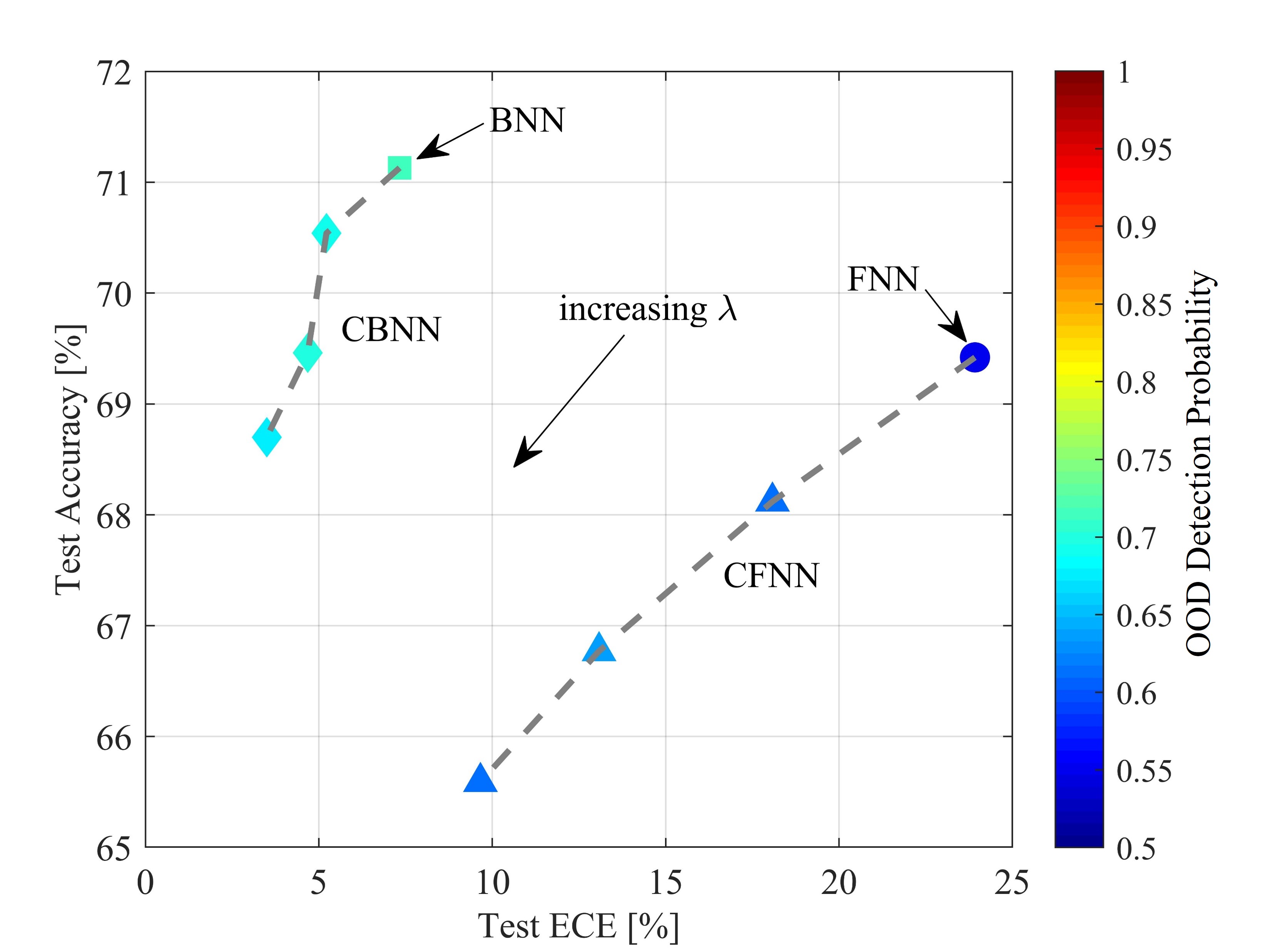}}
    \caption{Accuracy versus ECE obtained by changing the hyperparameter $\lambda$ on CIFAR-100 data set for FNN, CFNN (benchmark), BNN, and CBNN (ours), with OOD detection probability indicated by the marker's color. }
    \label{result:CA-pareto} 
    \vspace{-0.35cm}
\end{figure} 

Fig.~\ref{result:CA-pareto} demonstrates the trade-off between accuracy and ECE obtained by varying the hyperparameter $\lambda$ in (\ref{CA-FNN}) and (\ref{eq:CA-BNN_gen_recipe}). Note that with $\lambda=0$, one recovers FNN and BNN from CFNN and CBNN, respectively. Increasing the value of $\lambda$ is seen to decrease the ECE for both CFNN and CBNN and to gradually affect the accuracy, tracing a trade-off between ID calibration and accuracy. This figure also report on the OOD detection probability, which will be discussed later in this section.

\subsection{Can Confidence Minimization Improve OOD Detection without Affecting ID Performance?}

\begin{figure*} [htb] 
    \centering
    \centerline{\includegraphics[scale=0.5]{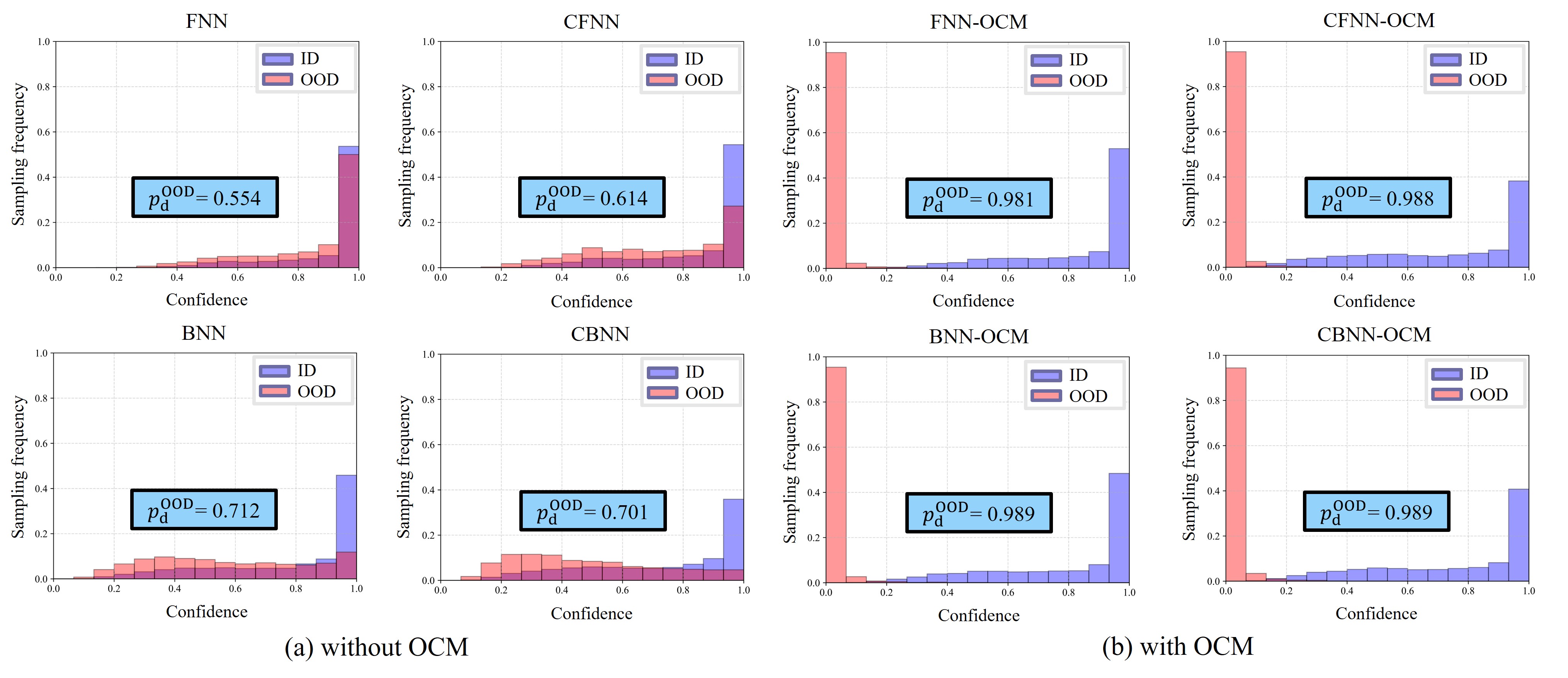}}
    \vspace{-0.5cm}
    \caption{Confidence histograms for ID and OOD data  drawn from CIFAR-100 and TinyImageNet data sets, respectively, for FNN, CFNN, FNN-OCM (benchmark), CFNN-OCM, BNN, CBNN, BNN-OCM, and CBNN-OCM (ours), with hyperparameter $\gamma = 0.5$ \cite{choi2023conservative}.}
    \label{result:OCM} 
\end{figure*}

To evaluate the performance in terms of OOD detection, we select ID data from the CIFAR-100 data set, which is also used for training, while OOD is selected from the TinyImageNet data set. As a reference, the color map in Fig.~\ref{result:CA-pareto} shows that the OOD detection probability (\ref{eq:OOD-detection-probability}) does not necessarily increase as the ID calibration performance improved, which is most notable for BNN. This highlights the importance of introducing mechanism tailored to enhancing OOD detection performance, such as OCM.

To evaluate the benefits of OCM, Fig.~\ref{result:OCM} plots the histograms of the confidence levels produced by different models for ID and OOD data. The more distinct the two distributions are, the larger the OOD detection probability (\ref{eq:OOD-detection-probability}) is \cite{choi2023conservative, polyanskiy2022information}. First, it is observed that BNN can improve OOD detection as compared to FNN, but the OOD confidence levels tend to be approximately uniformly distributed. Second, calibration-regularization, while improving ID calibration, does not help with OOD detection, as it focuses solely on ID performance. Finally, OCM can drastically enhance OOD detection performance for both FNN and BNN, with and without calibration-regularizers. In particular, thanks to OCM, the model produces low confidence levels on OOD inputs, enhancing the OOD detection probability (\ref{eq:OOD-detection-probability}).

\begin{figure} [htb] 
    \centering
    \centerline{\includegraphics[scale=0.35]{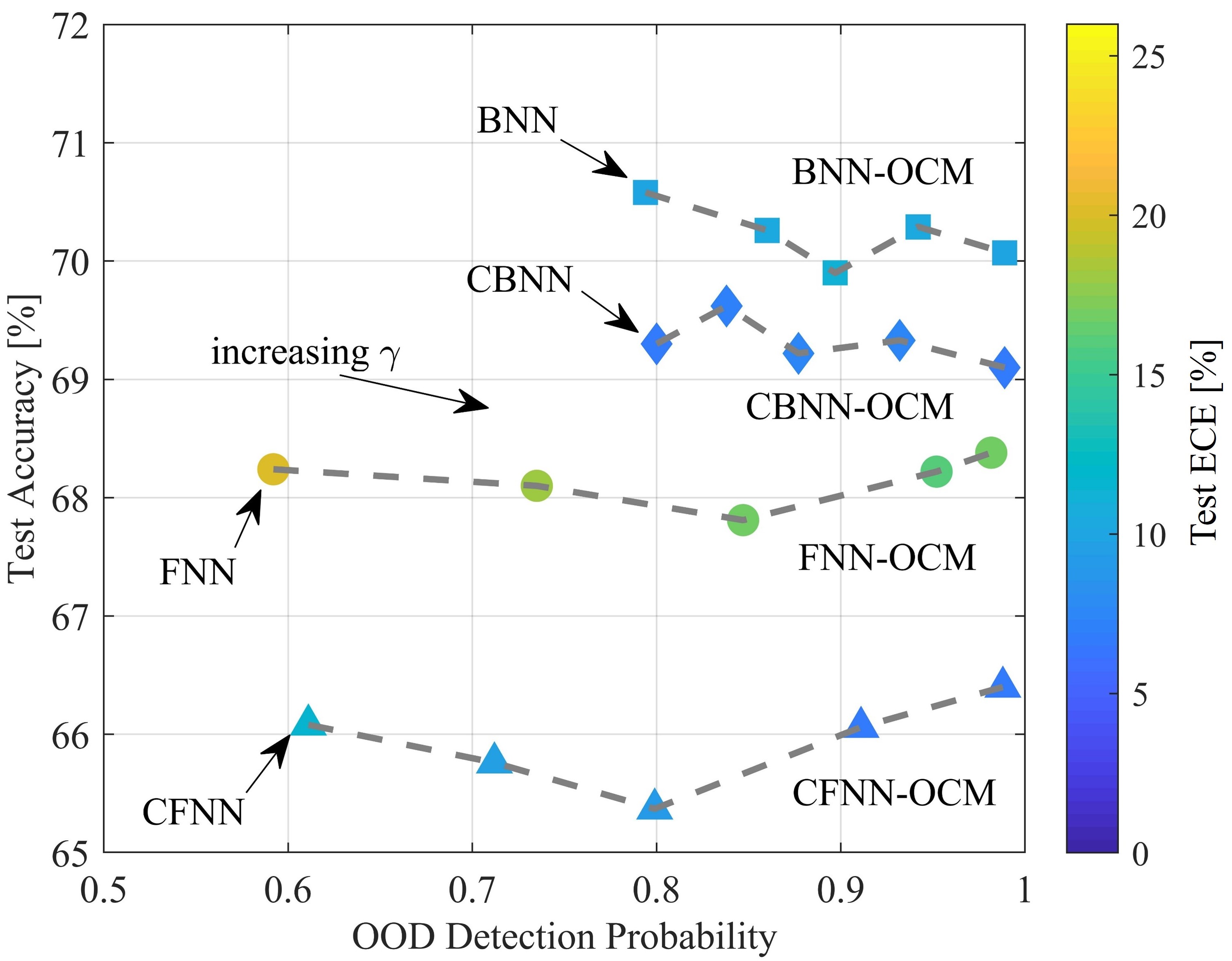}}
    \caption{Test accuracy versus OOD detection probability for FNN-OCM (benchmark), CFNN-OCM, BNN-OCM, and CBNN-OCM (ours), with test ECE as the marker's color. Note that hyperparameter $\gamma = 0$ recovers the original schemes without OCM regularizer.}\vspace{-0.7cm}
    \label{result:OCM-trade-off} 
\end{figure}

To further investigate the interplay between ID and OOD performance, Fig.~\ref{result:OCM-trade-off} plots the ID test accuracy versus the OOD detection probability by varying the hyperparameter $\gamma$ in (\ref{eq:CM}) and (\ref{eq:general_CM_gen_recipe}). BNN yields better test accuracy for any given OOD detection probability level, and adding calibration-regularizer improves ID calibration performance, but at the cost of a reduced ID accuracy for a given OOD detection probability level.

\subsection{Can Selective Calibration Compensate for the ID Performance Loss of OCM?}

So far, we have seen that there is generally a trade-off between ID and OOD performance. In particular, from Fig.~\ref{result:OCM-trade-off}, it was concluded that, for a given fixed level of OOD detection probability, calibration-regularized learning improved ID calibration at the cost of the ID accuracy for BNN-OCM. In this subsection, we ask whether selective calibration can ensure a synergistic use of OCM and calibration-aware regularization, guaranteeing that CBNN-OCM achieves the best ID and OOD performance levels. To address this question, we vary the ID coverage rate, i.e., the fraction of accepted ID samples, by changing the hyperparameter $\tau$ in (\ref{eq:scbnn-ocm}). The resulting ID and OOD performance is shown in Fig.~\ref{result:SEL-cal} for SBNN-OCM and SCBNN-OCM.

The figure demonstrates that, thanks to selective calibration, SCBNN-OCM outperforms SBNN-OCM in terms of all metrics, namely ID accuracy, ID calibration, and OOD detection probability, for a sufficiently low ID coverage rate, here smaller than 50\%. We thus conclude that, by selecting examples with sufficiently good ID calibration performance, SCBNN-OCM can fully benefit from calibration-aware regularization, enhancing ID performance, as well as from OCM, improving OOD detection performance. That said, the need to accept a sufficiently small number of inputs to ensure this result highlights the challenges of guaranteeing both ID and OOD performance levels.

\subsection{Memory Requirements and Computational Cost} Finally, we analyze the memory requirements and computational cost of the different schemes studied in this paper. To this end, we focus on the setting adopted for the experiments reported in Fig.~\ref{result:CA-RD}, Fig.~\ref{result:OCM}, and Fig.~\ref{result:SEL-cal}. We evaluate the number of model parameters and the number of floating-point operations (FLOPs) during inference. The number of FLOPs is reported normalized by the size of the ensemble, and is evaluated using a standard profiler \cite{chitale2023task}. The results, reported in Table~\ref{tab:overhead}, can be used to determine the trade-off between the performance improvements observed in Fig.~\ref{result:CA-RD}, Fig.~\ref{result:OCM}, and Fig.~\ref{result:SEL-cal} and the memory and computational costs. The overall number of FLOPs can be obtained from Table~\ref{tab:overhead} by multiplying by the ensemble size, which is set to $20$ for inference and to $1$ for training. Furthermore, the number of training FLOPs per iteration must account for the cost of gradient evaluation, which is typically approximated as $2$-$5$ times the inference cost \cite{goodfellow2016deep, kaplan2020scaling}.

\begin{table}[ht]
  \centering
  \caption{Number of parameters and number of inference FLOPs per ensemble size (FLOPs are calculated with image size $32 \times 32$).}
  \label{tab:overhead}
  \begin{tabular}{@{} l l r r r @{}}
    \toprule
    Model                          & Parameters  &   FLOPs/ensemble size\\
    \midrule
    FNN   &  2.25M        &  0.3G      \\ 
    CFNN   & 2.25M         & 0.3G         \\
    BNN               & 4.5M         & 0.34G           \\
    CBNN            &4.5M         & 0.34G    \\
    \midrule
    FNN-OCM    &  2.25M         & 0.3G        \\
    CFNN-OCM   &  2.25M         & 0.3G       \\
    BNN-OCM                & 4.5M         &0.34G       \\
    CBNN-OCM             & 4.5M        & 0.34G       \\
    \midrule
    Selector   &  0.005M          &  0.001G       \\ 
    \bottomrule
  \end{tabular}
\end{table}

\begin{figure*} [htb] 
    \centering
    \centerline{\includegraphics[scale=0.385]{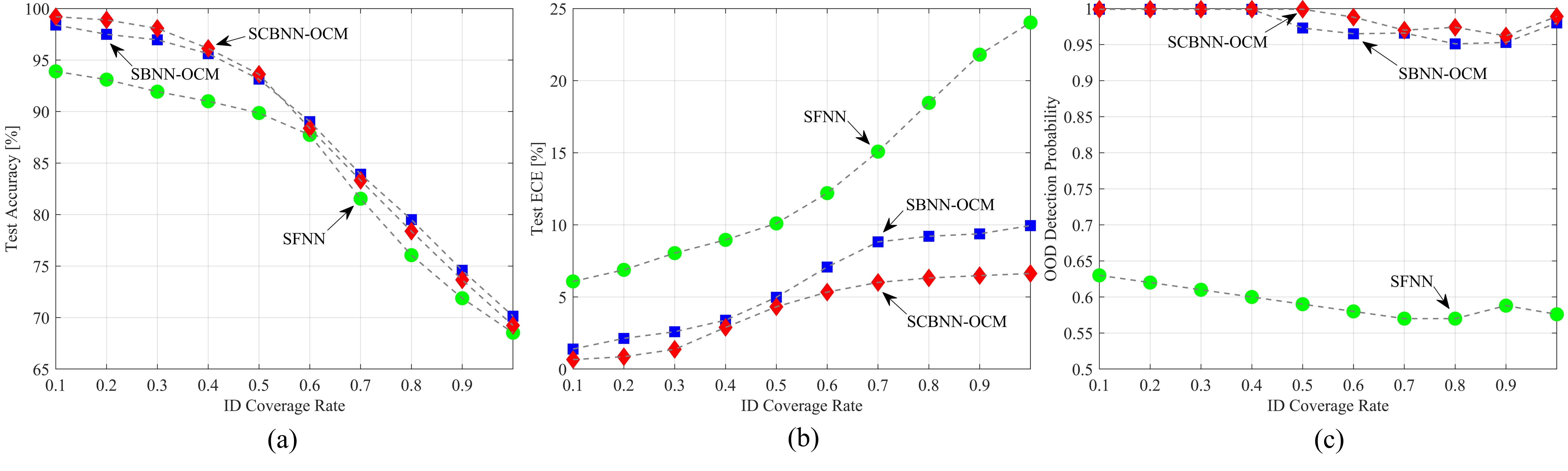}}
    \caption{Test accuracy, ECE, and OOD detection probability versus ID coverage rate for SFNN (benchmark), SBNN-OCM and SCBNN-OCM (ours).}
    \label{result:SEL-cal} 
\end{figure*}
\vspace{-0.1cm}

\vspace{-0.2cm}
\section{Conclusion} \label{sec:conclusion}
In this paper, we have proposed SCBNN-OCM, a general framework that enhances variational inference-based Bayesian learning to target \emph{both} ID and OOD calibration. To improve ID calibration, we have introduced a regularizer based on the calibration error, while OOD calibration is enhanced by means of data augmentation based on confidence minimization. In order to facilitate the synergistic use of calibration-aware regularization and OCM regularization, we have finally introduced a selective calibration strategy that rejects examples that are likely to have poor ID calibration performance. Numerical results have illustrated the challenges in ensuring both ID and OOD performance, as schemes designed for ID calibration may end up hurting OOD calibration, and vice versa. That said, thanks to selective calibration, SCBNN-OCM was observed to achieve the best ID and OOD performance as compared to the other benchmarks as long as one allows for a sufficiently large number of rejected samples. For instance, as compared to standard FNN, SCBNN-OCM achieves a $25\%$ improvement in accuracy, a $20\%$ drop in ECE, and a $50\%$ improvement in OOD detection probability, at the cost of reduced ID coverage rate of around $50\%$. Interesting directions for future work include extending the proposed framework to likelihood-free, simulation-based inference \cite{falkiewicz2024calibrating}, to continual learning \cite{li2024calibration}, and to large language models \cite{detommaso2024multicalibration}. 
\bibliographystyle{IEEEtran}
\bibliography{refs}
\newpage
\clearpage

\begin{center}
    {\Large \textbf{Supplementary Material for ``Calibrating-Bayesian Learning via Regularization, Confidence Minimization, and Selective Inference''}}\\[1em]
\end{center}

\begin{abstract}
    This supplementary document consists of four sections. Ablation studies on ID calibration, OOD detection, and selective calibration are conducted to illustrate the impact of hyperparameters, i.e., weight of regularize, in Sec. \ref{sec1}, Sec. \ref{sec2}, and Sec. \ref{sec3}, respectively. In Sec. \ref{sec4}, to showcase the general performance gains of the proposed framework, we present the additional evaluations on a different backbone. We refer to appendix for all the experimental details.
\end{abstract}

\section{ID Calibration Ablation} \label{sec1}
\subsection{Impact of Calibration-Based Regularizer Weight}
\begin{figure} [htb] 
    \centering
    \centerline{\includegraphics[scale=0.036]{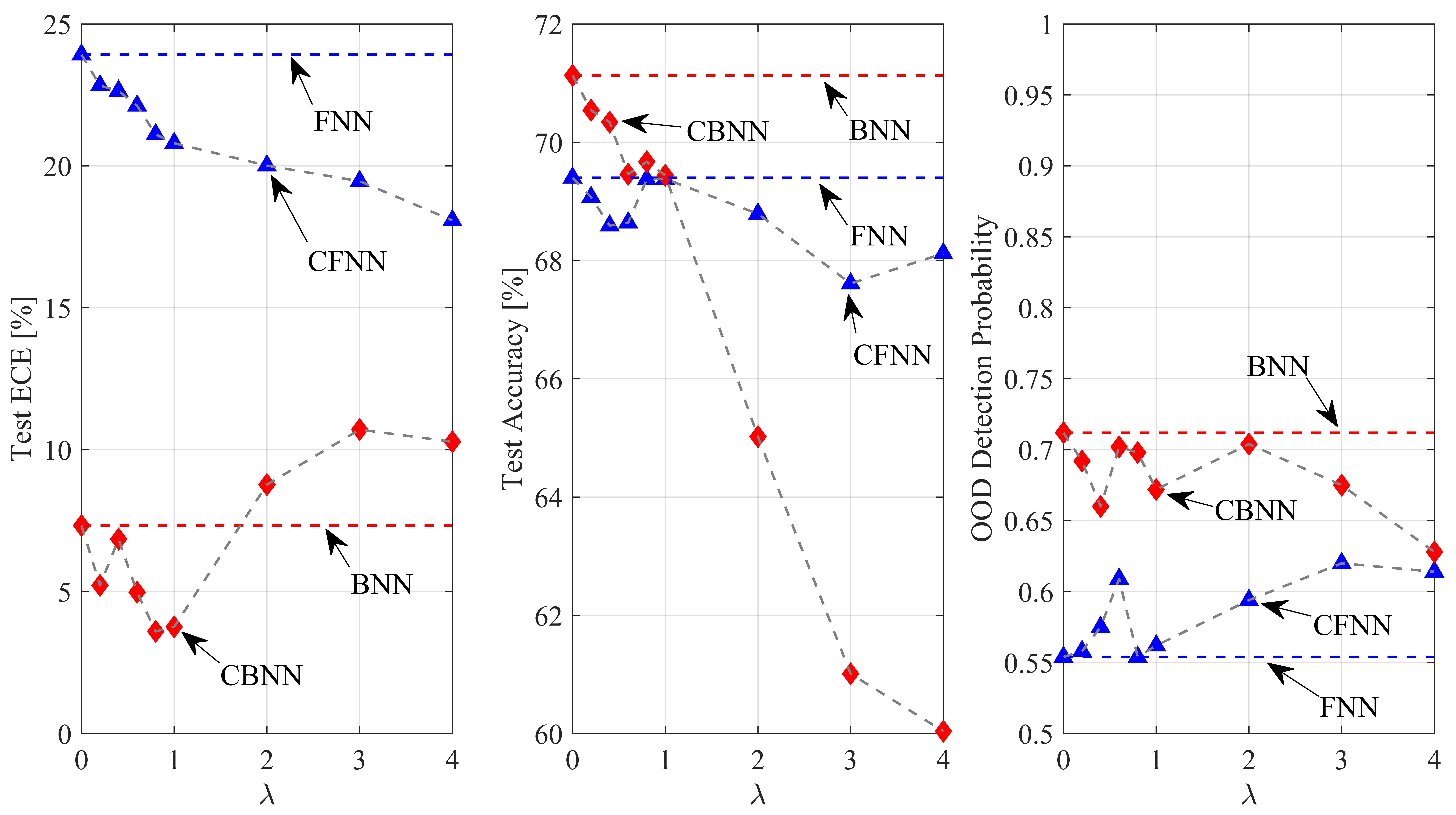}}
    \caption{ECE, accuracy and OOD detection probability as a function of hyperparameter $\lambda$ on CIFAR-100 (ID) and TinyImageNet (OOD) data set for FNN, BNN, CFNN and CBNN.}
    \label{ablation:CA_1} 
\end{figure}

    The impact of the calibration-based regularizer hyperparameter, $\lambda$ in (9) and (17), for ID calibration is analyzed in Fig. \ref{ablation:CA_1} under the same conditions as for Fig. 7 in the main text. The figure confirms that the best performance for CFNN and CBNN on CIFAR-100 (ID) and TinyImageNet (OOD) data sets is achieved for $\lambda = 4$ and $\lambda = 0.8$, respectively. CBNN scheme is seems to be more sensitive to the selection of hyperparameters $\lambda$, possibly because of the interplay with the choice of the prior distribution in the Bayesian learning objective (17). That said, CBNN outperforms BNN in terms of ECE for a wide range of values of hyperparameter $\lambda$.

\section{OOD Detection Ablation} \label{sec2}
\subsection{Impact of OCM Regularizer Weight}
\begin{figure} [htb] 
    \centering
    \centerline{\includegraphics[scale=0.04]{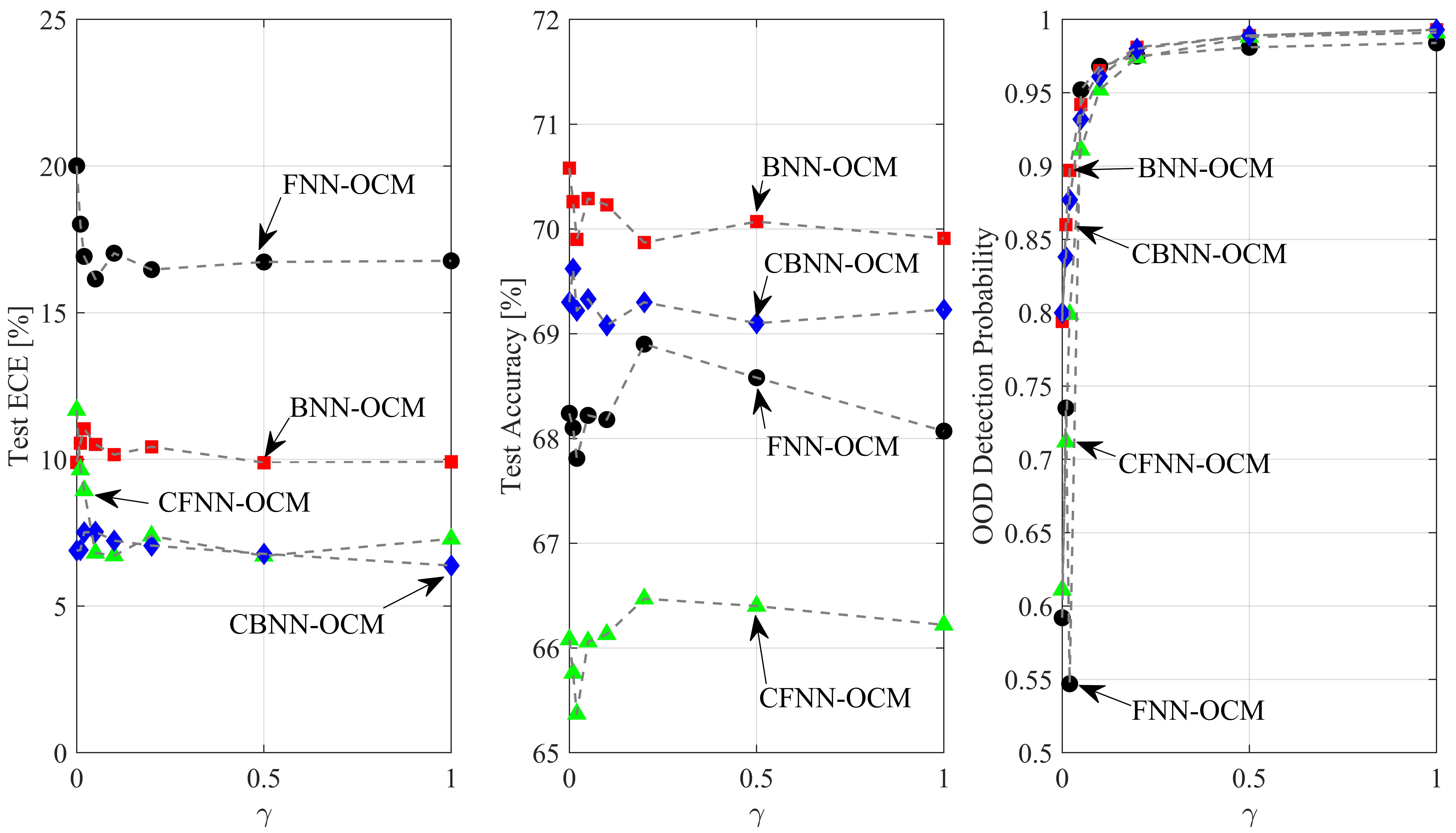}}
    \caption{ECE, accuracy and OOD detection probability as a function of hyperparameter $\gamma$ on CIFAR-100 (ID) and TinyImageNet (OOD) data set for FNN-OCM, BNN-OCM, CFNN-OCM and CBNN-OCM.}
    \label{ablation:CM_1} 
\end{figure}

    Fig. \ref{ablation:CM_1} presents an ablation study for the OCM regularizer weight,  $\gamma$ in (21) and (25), for the same setting as in the previous subsection. The ECE, test accuracy, and OOD detection probability are relatively stable when $\gamma > 0.1$, which implies that our default choice of $\gamma = 0.5$, also reported in \cite{choi2023conservative}, is well suited for all schemes.

\section{Selective Calibration Ablation} \label{sec3}
\subsection{Impact of Selector Regularizer Weight}
\begin{figure} [htb] 
    \centering
    \centerline{\includegraphics[scale=0.04]{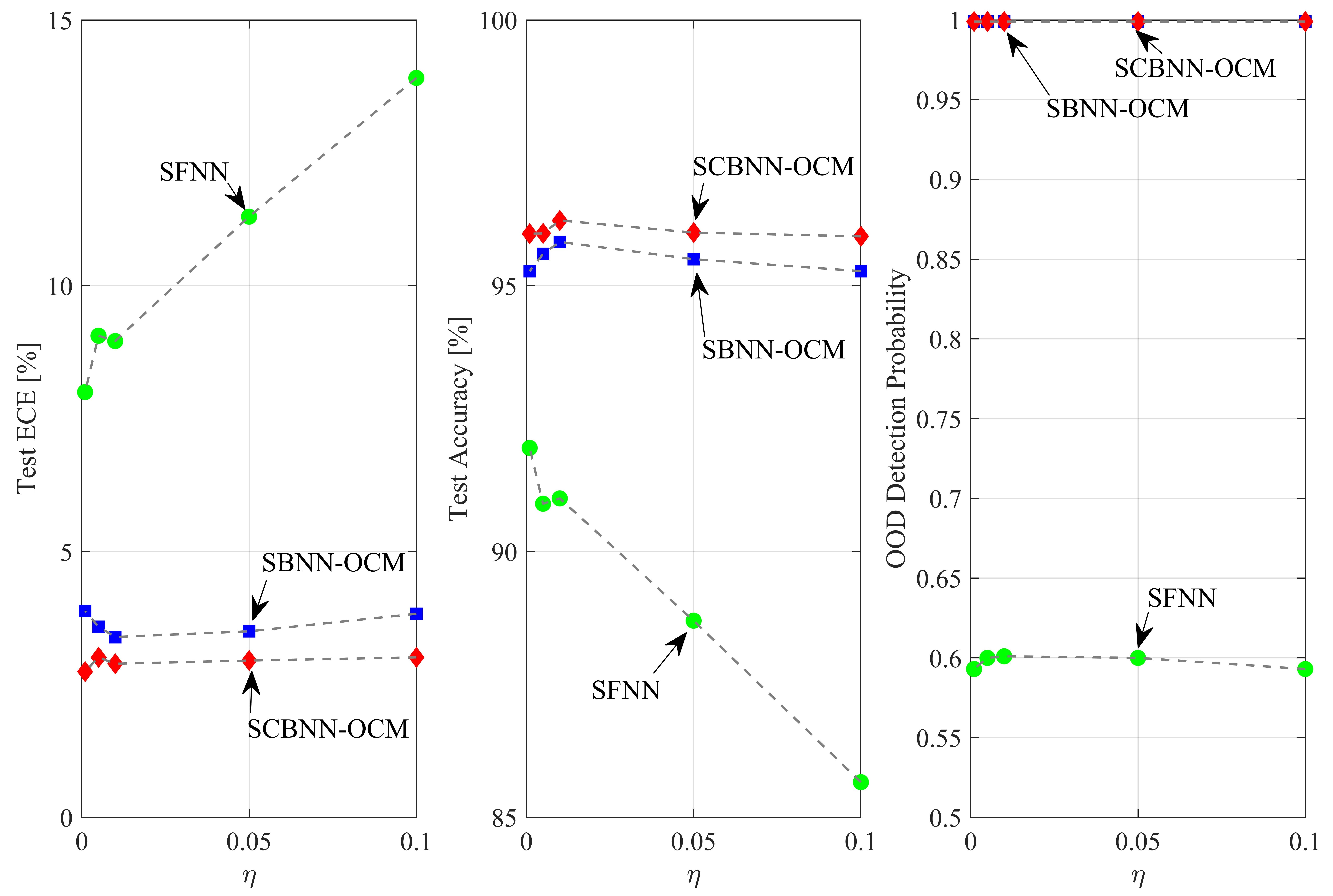}}
    \caption{ECE, accuracy and OOD detection probability as a function of hyperparameter $\eta$ on CIFAR-100 (ID) and TinyImageNet (OOD) data set for SFNN, SBNN-OCM, and SCBNN-OCM with ID coverage rate as $40\%$.}
    \label{ablation:SEL_1} 
\end{figure}

    By varying the selector regularizer weight, $\eta$ in (33) and (39), Fig. \ref{ablation:SEL_1} shows that the proposed Bayesian methods have a better performance in terms of all metrics as compared to SFNN, and are more robust to the choice of hyperparameter $\eta$. To maintain consistency with \cite{fisch2022calibrated}, we set $\eta = 0.01$ as the default choice for $\eta$.

\section{Backbone Study} \label{sec4}
    In order to showcase that the gain of SCBNN-OCM is not restricted to a particular neural network architecture or to a specific ID/OOD data set, we present additional evaluations on a different backbone, namely ResNet-18, and on different data sets by considering CIFAR-10 as the ID and LSUN resized as the OOD data set. For all the experiments in this section, we use the same settings as the experimental results in the main text.
\subsection{Impact of the Calibration-Based Regularizer}
\begin{figure} [htb] 
    \centering
    \centerline{\includegraphics[scale=0.32]{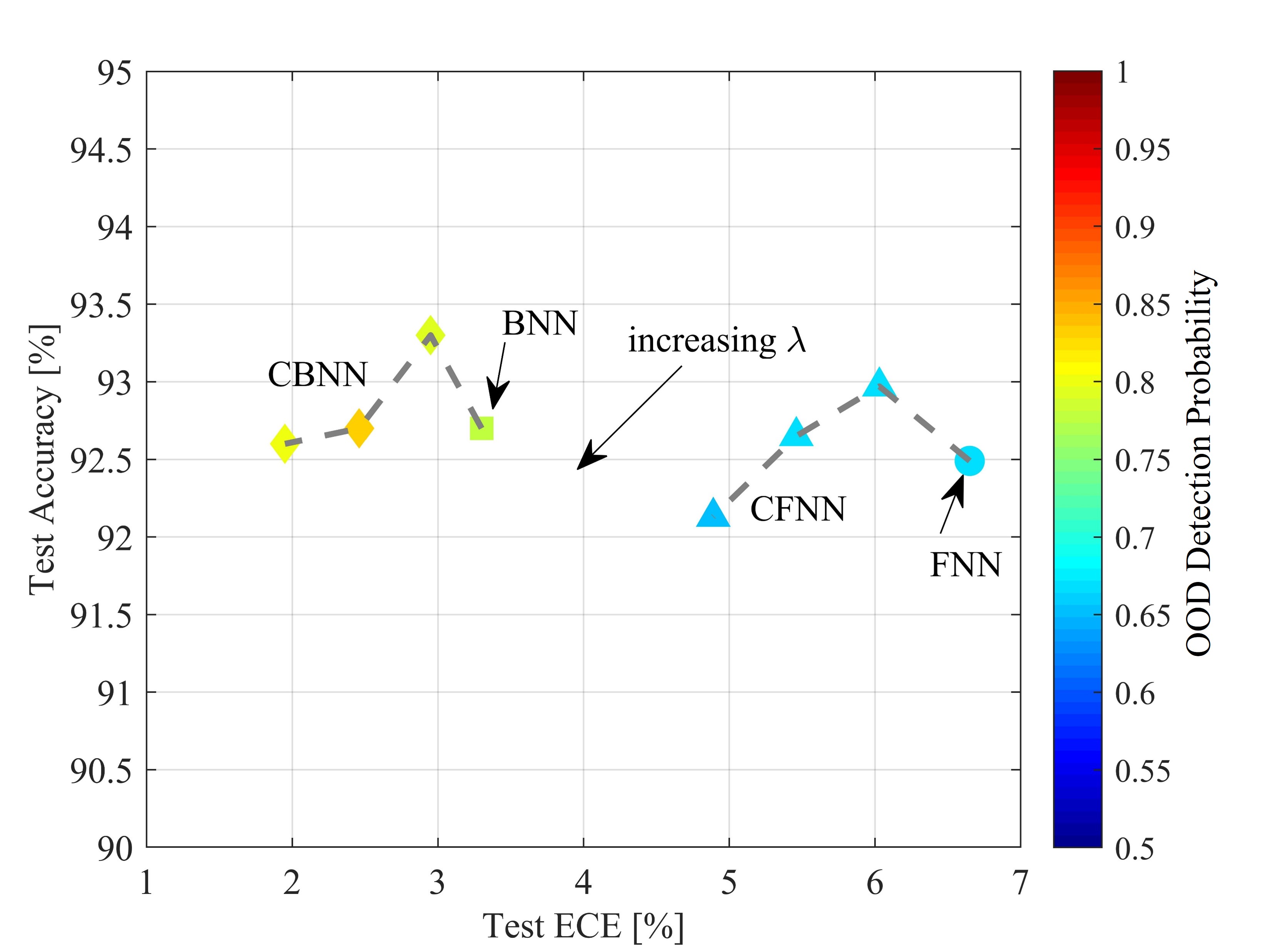} }
    \caption{Accuracy versus ECE obtained by changing the hyperparameter $\lambda$ on CIFAR-10 data set for FNN, CFNN (benchmark), BNN, and CBNN (ours), with OOD detection probability indicated by the marker's color.} 
    \label{result:backbone_1}  
\end{figure}

    The impact of the calibration-based regularization hyperparameter for ID calibration, $\lambda$ in (9) and (17), is analyzed in Fig. \ref{result:backbone_1}. The figure confirms that the best performance for CFNN and CBNN on CIFAR-10 (ID) and LSUN (OOD) data sets is achieved for $\lambda = 10$ and $\lambda = 4$, respectively. Also, increasing the value of $\lambda$ is seen to decrease the ECE for both CFNN and CBNN, with similar test accuracy.
\subsection{Impact of the OCM Regularizer}
\begin{figure} [htb] 
    \centering
    \centerline{\includegraphics[scale=0.32]{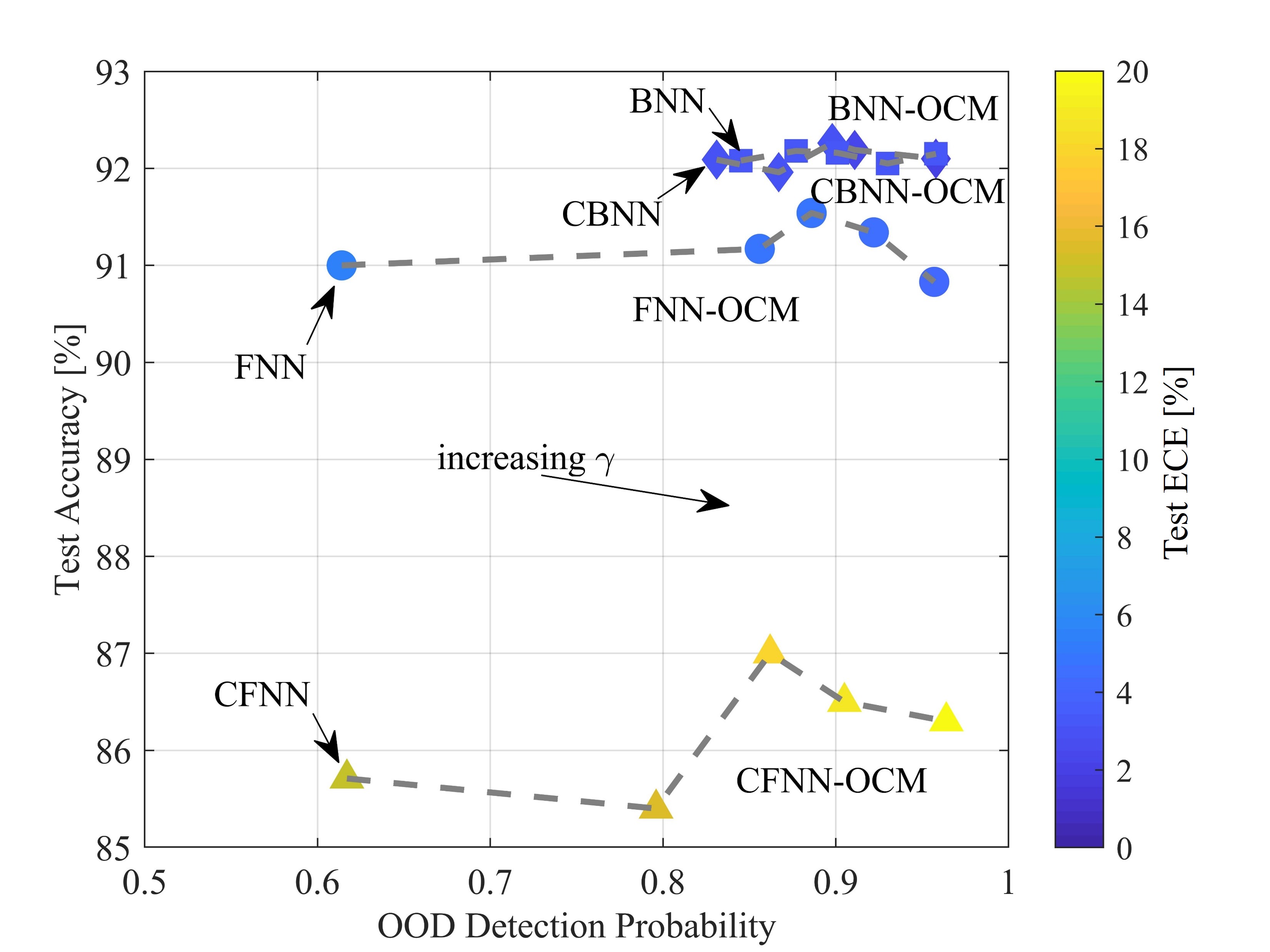} }
    \caption{Test accuracy versus OOD detection probability on CIFAR-10 (ID) and LSUN (OOD) for FNN-OCM (benchmark), CFNN-OCM, BNN-OCM, and CBNN-OCM (ours), with test ECE as the marker's color. Note that hyperparameter $\gamma = 0$ recovers the original schemes without OCM regularizer.} 
    \label{result:backbone_2}  
\end{figure}

    To investigate the impact of OCM regularizer on ID and OOD performance, Fig. \ref{result:backbone_2} plots the ID and OOD performance by varying the OCM hyperparameter $\gamma$ in (21) and (25). Note that the deterioration of the ID performance, especially for CFNN-OCM, shows the side effects of introducing OCM, although it entails OOD detection probability improvement. Additionally, the proposed CBNN-OCM better balances the ID and OOD performance levels as compared to CFNN-OCM.

\subsection{Impact of Selective Calibration}
    To investigate the impact of selective calibration, Fig. \ref{result:SEL-cal} plots the ID and the OOD performance for different ID coverage rates ranging from $0.1$ to $1$. The figure shows that SCBNN-OCM outperforms SBNN-OCM in terms of ECE and OOD detection probability in different ID coverage rate regimes, while attaining similar test accuracy.
    
    As compared to the Fig. 10 in the main text, SCBNN-OCM is seen to achieve better ID and OOD performance with relatively small reduction in terms of ID coverage rate. For instance, with an ID coverage rate of $0.7$, the ID accuracy is around $99\%$, the ECE is lower than $1$, and the OOD detection probability is nearly $1$.

\begin{figure*} [htb] 
    \centering
    \centerline{\includegraphics[scale=0.36]{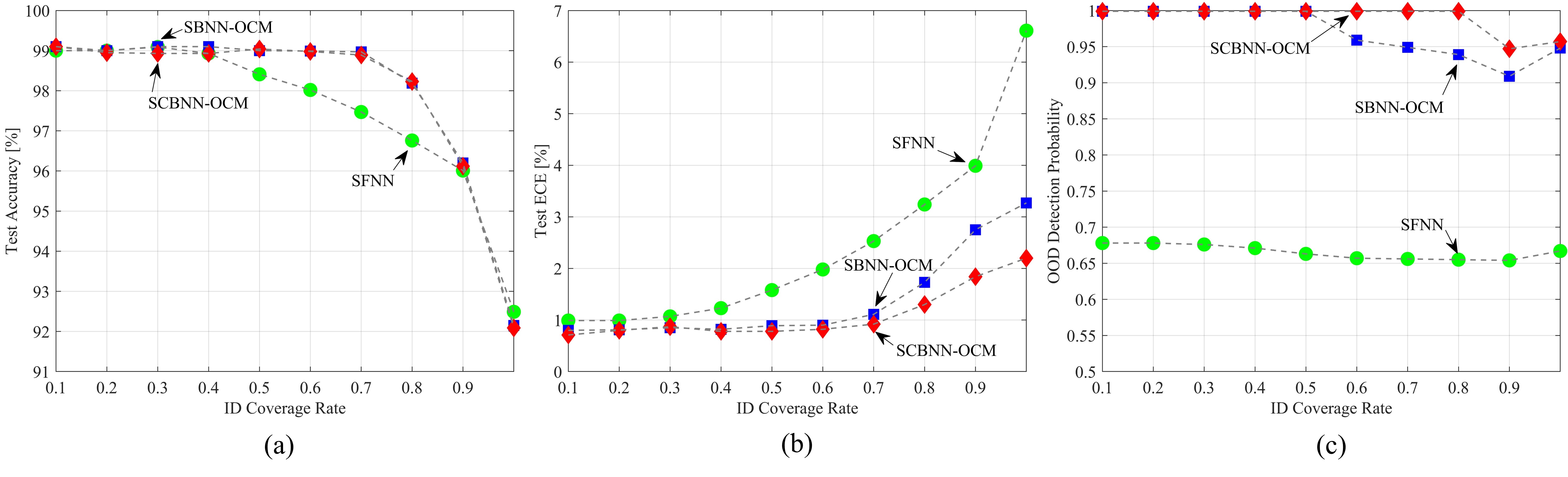}}
    \caption{Test accuracy, ECE, and OOD detection probability versus ID coverage rate on CIFAR-10 (ID) and LSUN (OOD) for SFNN (benchmark), SBNN-OCM and SCBNN-OCM (ours).}
    \label{result:SEL-cal} 
\end{figure*}

\appendix 

\subsection{Non-Parametric Outlier Score Vectors} \label{appdendix:Non-parametric score estimator}
In this subsection, we specify the details for non-parametric outlier score vector $ s (x|\theta^\text{FNN})$ described in Sec.~IV-B2 of the main text, which can be implemented by following the sklearn package in Python. As mentioned in the main text, the first entry is given by (35) with the Gaussian kernel $\kappa \left( || z-z^{\text{tr}}_{i} || \right) = \exp \left( - || z - z_i^{\text{tr}} ||^2/h \right)$ given the bandwidth parameter $h>0$.
   
The isolation forest score for $x$, $s_2 (x|\theta^\text{FNN})$, is obtained by evaluating the \emph{depth} of $z$ in the \emph{isolation forest}, which is a collection of $T$ binary trees  in which each binary tree is constructed so that every element in $\{ z_i^\text{tr} \}_{i=1}^{|\mathcal{X}^\text{tr}|} \cup \{z\}$ is assigned to a leaf node. Specifically, the isolation forest score is evaluated as \cite{liu2008isolation}
\begin{align}
   s_2 (x|\theta^\text{FNN}) = 2^{-\frac{\sum_{t=1}^T h_t(z)}{T\cdot c}}, \label{eq:isolation_forest}
\end{align}
where $h_t(z)$ is the depth of $z$ in the $t$-th binary tree given a normalization constant $c$. 

The one-class support vector machine (SVM) score for $x$, $s_3 (x|\theta^\text{FNN})$, can be expressed as \cite{scholkopf2001estimating}
\begin{align}
   s_3 (x|\theta^\text{FNN}) = \ \sum_{i=1}^{|\mathcal{X}^{\text{tr}}|} \alpha_i \kappa \left( || z-z^{\text{tr}}_{i} || \right) - \rho ,
\end{align}
in which the coefficients $\alpha_i \geq 0$ for $i=1,...,|\mathcal{X}^\text{tr}|$, satisfying $\sum_{i=1}^{|\mathcal{X}^\text{tr}|} \alpha_i=1$, are optimized along with the offset parameter $\rho \in \mathbb{R}$ in order to separate the training data $\{ z_i^\text{tr} \}_{i=1}^{|\mathcal{X}^\text{tr}|}$ from the origin in the feature space associated with the kernel $\kappa(\cdot)$ \cite[Sec. 5]{scholkopf2001estimating}.    

Finally, $k$-nearest neighbor distance for $x$, $s_4 (x|\theta^\text{FNN})$, can be written as \cite{loftsgaarden1965nonparametric}
\begin{align}
    s_4 (x|\theta^\text{FNN}) = || z - z_i^{\text{tr}} ||_{(k)},
\end{align}
where $|| z - z_i^{\text{tr}} ||_{(k)}$ is the $k$-th smallest value in the set of distances $\{|| z - z_i^{\text{tr}} ||\}_{i=1}^{|\mathcal{X}^{\text{tr}}|}$.

\subsection{Experiment Details}  \label{appendix:ID-calibration-settings}
In this subsection, we specify the implementation details for the experimental results in Sec.~V.  All the experiments are implemented by PyTorch \cite{paszke2019pytorch} and run over a GPU server with single NVIDIA A100 card.
\subsubsection{Architecture and Training Details for Calibration-Regularized Learning}\label{subsec:cal}\text{ } 

\emph{Architecture}: For all the experiments related to calibration-regularized learning, we adopt the WideResNet-40-2 architecture \cite{zagoruyko2016wide}.

\emph{Hyperparameters}: For fair comparison, we use the same training policy for both frequentist and Bayesian learning. Specifically, we use the SGD optimizer with momentum factor $0.9$ and train the model during $100$ epochs. We also decrease the learning rate by dividing its value by a factor of $5$ for every $30$ epochs with initial learning rate $0.1$ in a manner similar to \cite{choi2023conservative}.
The minibatch size used for single SGD update is set to $128$.  For Bayesian learning, we set as  prior $p(\theta)$ the Gaussian distribution that has zero-mean vector with diagonal covariance matrix having each element as $0.001$. The hyperparamter $\beta$ for the free energy (11) in the main text is set to $0.00035$. The ensemble size for Bayesian learning during training is set to $1$, while the ensemble size during testing is set to $20$.
In a manner similar to \cite{yoon2023esd}, the calibration-based regularizer $\lambda$ in (9) and (17) in the main text is chosen as the value in set $\{0.2, 0.4, 0.6, 0.8, 1.0, 2.0, 3.0, ..., 10.0\}$ that achieves the lowest ECE, while preserving the accuracy drop no larger than $1.5\%$ as compared to a setting with $\lambda=0$. The corresponding accuracy and ECE are evaluated based on the validation data set $\mathcal{D}^\text{val}$, and the hyparparameter $\lambda$ for CFNN and for CBNN are chosen independently, which result in the values of $4$ and $0.8$ respectively in our experiments. With these choices, CFNN and CBNN can achieve lower ECE with acceptable accuracy drops. We use weighted MMCE regularizer for calibration-aware regularization term $\mathcal{E}(\theta|\mathcal{D}^\text{tr})$, which is defined as \cite{kumar2018trainable}
\begin{align} 
     \mathcal{E}(\theta|\mathcal{D}^\text{tr}) &= \Bigg(\sum_{i,j:{c}_i = {c}_j = 0} \frac{{r}_i {r}_j \kappa({r}_i , {r}_j)}{(|\mathcal{D}^\text{tr}|-n_c)(|\mathcal{D}^\text{tr}|-n_c)}  \nonumber \\ &+ \sum_{i,j:{c}_i = {c}_j = 1} \frac{(1-{r}_i) (1-{r}_j) \kappa({r}_i , {r}_j)}{n_c ^2}
      \nonumber \\ &-2 \sum_{i,j:{c}_i = 1,  {c}_j = 0} \frac{(1-{r}_i) {r}_j \kappa({r}_i , {r}_j)}{(|\mathcal{D}^\text{tr}|-n_c)n_c} \Bigg)^{\frac{1}{2}},
\end{align}
where the kernel function is $ \kappa(r_i, r_j) = \exp (-||r_i - r_j||/0.4) $, and $n_c$ is the number of correct examples, i.e., $n_c = \sum_{i=1}^{|\mathcal{D}^\text{tr}|} c_i.$

\emph{Data set split and augmentations}: As mentioned in Sec.~V-A, we choose CIFAR-100 data set \cite{krizhevsky2010cifar} for the ID samples, which is a data set composed of $60,000$ images each with label information chosen among $100$ different classes. In particular, CIFAR-100 splits the data set into two parts: $50,000$ for training and $10,000$ for testing; and we further split the training data set into $45,000$ examples and $5,000$ examples to define the training data set $\mathcal{D}^\text{tr}$ and the validation data set $\mathcal{D}^\text{val}$. We adopt the standard random flip and random crop augmentations provided by Pytorch \cite{paszke2019pytorch} during the training process.

\subsubsection{Architecture and Training Details for OOD Detection}  \label{appendix:OOD-detection-settings} \text{ }
\emph{Architecture}: Since we use the same predictor for OOD detection, the corresponding architecture remains the same, i.e., WideResNet-40-2, as described above.

\emph{Hyperparameters}: Following the original OCM paper \cite{choi2023conservative}, OOD confidence minimization (21) and (25) in the main text is carried out by fine-tuning based on the corresponding pre-trained models, e.g., CBNN-OCM is obtained via fine-tuning with the OCM-regularized training loss given the pre-trained CBNN. Uncertainty data set $\mathcal{D}^\text{u}$ is constructed by randomly choosing $6,000$ input data from TinyImageNet. During fine-tuning, SGD optimizer with momentum factor of $0.9$ is adopted during $10$ epochs, each consisting 282 iterations. The initial learning rate is set to $0.001$, and we update the learning rate for each iteration by following \cite{hendrycks2019oe}. Specifically, during SGD training, for each epoch, we first sample 9,000 examples from $\mathcal{D}^\text{tr}$ and evaluate the ID-related training measures for each SGD update via sampling 32 examples (minibatch size being 32) without replacement among the 9,000 examples; for the OOD-related training measures, we sample 64 examples (minibatch size being 64) among $\mathcal{D}^\text{u}$ with replacement. The hyperparameter of OCM regularizer $\gamma$ in (21) and (25) in the main text is chosen as in reference \cite{choi2023conservative}, i.e, $\gamma = 0.5$, which increases OOD detection probability nearly to $1$ for all OCM schemes. Other hyperparameters are the same as calibration-regularized learning described in the previous subsection.

\emph{Data set split and augmentations}: TinyImageNet data set \cite{liang2017principled} contains $10,000$ images that corresponds to different $200$ classes. We split the input data of TinyImageNet data set into two parts, $6,000$ and $4,000$, and use them for the uncertainty data set $\mathcal{D}^\text{u}$ and for the OOD test data set, respectively. During fine-tuning, we apply the same standard random flip and random crop augmentations to both $\mathcal{D}^\text{tr}$ and $\mathcal{D}^\text{u}$.

\subsubsection{Architecture and Training Details for Selective Calibration}  \label{appendix:selector-settings} \text{ }

\emph{Architecture}: For the selector implementation, we use a 3-layer feed-forward neural network with 64 neurons in each hidden layer, activated by ReLU. 

\emph{Hyperparameters}: For selector training (33) and  (39) in the main text, we use Adam optimizer, with learning rate $0.001$ and weight decay coefficient $10^{-5}$. We train the model for $5$ epochs,  each epoch consisting $50,000$ iterations, and each iteration samples $32$ examples (minibatch size being $32$) among $\mathcal{D}^\text{val}$. Since the proposed SCBNN-OCM is more robust to the choives of $\eta$, the hyperparameter $\eta$ in (33) and  (39) in the main text is set to $0.01$ as reported in \cite{fisch2022calibrated}, and we set the kernel function in (40) in the main text as $ \kappa(r_i, r_j) = \exp (-||r_i - r_j||/0.2) $. We use the same ensemble size for training and testing as in calibration-regularized learning.

\emph{Data set split and augmentations}: As described above, $\mathcal{D}^\text{val}$ has $5,000$ examples obtained from CIFAR-100 data set. We utilize the standard random flip and random crop augmentations during selector training.

\end{document}